\documentclass[11pt]{article}

\usepackage[preprint]{acl}

\usepackage{times}
\usepackage{latexsym}

\usepackage[T1]{fontenc}

\usepackage[utf8]{inputenc}

\usepackage{microtype}

\usepackage{inconsolata}

\usepackage{graphicx}

\usepackage{hyperref}
\usepackage{url}
\usepackage{booktabs}
\usepackage{makecell}
\usepackage{multirow}
\usepackage{amsmath}
\usepackage{amssymb}

\definecolor{darkblue}{rgb}{0, 0, 0.5}
\hypersetup{colorlinks=true, citecolor=darkblue, linkcolor=darkblue, urlcolor=darkblue}

\title{Towards Better Aligned LLM Judges for\\Radiology Report Evaluation across Modalities and Anatomies}

\author{Federica Bologna\footnote{Corresponding authors: fb265@cornell.edu, jcorbeil@microsoft.com} \\
Cornell University\\
Ithaca, NY, USA \\
\And
Jean-Philippe Corbeil\footnotemark[1] \\
Microsoft Healthcare \& Life Sciences \\
Montréal, QC, Canada \\
\AND
Matthew Wilkens \\
Cornell University\\
Ithaca, NY \\
\And
Asma Ben Abacha \\
Microsoft Healthcare \& Life Sciences \\
Redmond, WA, USA\\
}

\begin{document}
\maketitle

\begin{abstract}
Current work on radiology report evaluation has focused primarily on LLM-based metrics for chest X-rays (CXR), leaving open whether such approaches remain robust across other modalities and anatomies, and which model and prompt configurations are best suited to serve as LLM judges. We present a comprehensive evaluation of LLM judges for radiology report evaluation across diverse model families, prompting strategies, and adaptation methods, leveraging two expert-annotated datasets that span multiple imaging modalities and anatomical regions. Specifically, we compare four existing metrics (RadFact, GREEN, FineRadScore, and CRIMSON) with our proposed metric, VERT, across open- and closed-source models of varying sizes and reasoning capabilities. Among existing metrics, GREEN proves the most robust, outperforming more recent CXR-specific approaches. VERT further improves Kendall's correlation with radiologist judgment by up to 26\% over GREEN, while ensembling and lightweight fine-tuning of Qwen3-30B yield gains of up to 9.9\% and 25\%, respectively, of which the latter also has a latency reduction of 37.2×. A systematic error analysis further shows that, while these metrics correlate with expert ratings overall, they tend to miscount errors such as omissions. Overall, our results show that carefully selected general-purpose LLM judges provide strong radiology report evaluation across modalities, with lightweight adaptation offering complementary gains when annotated data is available.
\end{abstract}

\section{Introduction}

The evaluation of radiology reports has largely followed two separate lines: fine-tuning small models on LLM-generated labels and designing LLM-based metrics centered on error detection \citep{huang2024fineradscore,ostmeier2024green,zambrano2025chexprompt,baharoon2026crimson}. Beyond concentrating almost exclusively on chest radiographs (CXR), these efforts rest on a narrow empirical basis---most rely on ReXVal \citep{yu2023radcliq,PhysioNet-rexval-dataset-1.0.0}, a 50-report benchmark whose reliability has recently been questioned \citep{li2025reevalmed}. Finetuned metrics in particular tend to degrade beyond CXR \citep{cho2025crepe}, as differences in modality, anatomical focus, and reporting conventions are hard to generalize across. Thus, a metric optimized for chest reports may miss nuanced findings in domains such as MRI or CT.

\begin{figure}
    \centering
    \includegraphics[width=\linewidth]{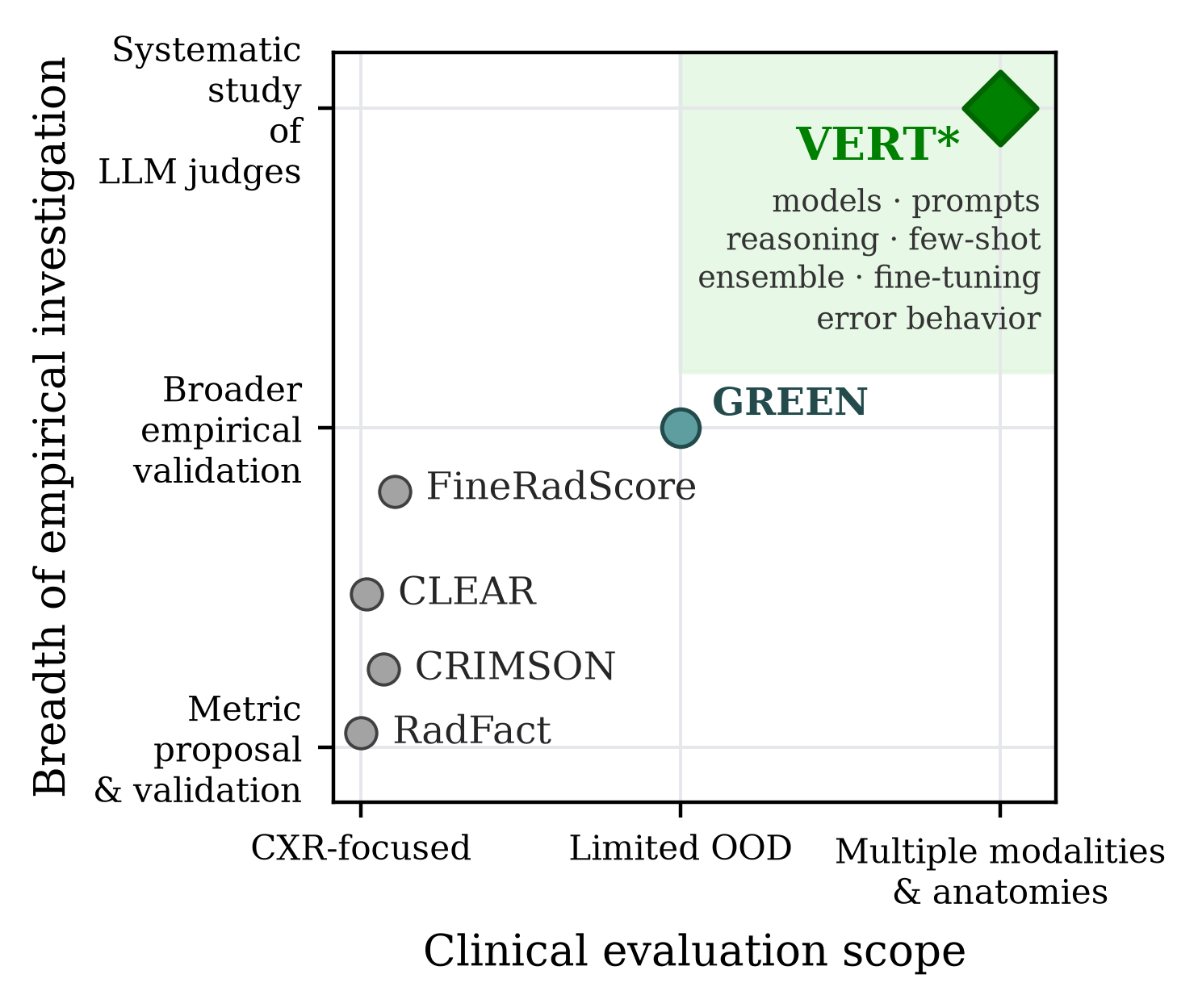}
    \caption{Clinical scope and empirical breadth of prior work and current paper noted VERT*, emphasizing the complete study and not only our VERT metric. \textit{The metric positions are for illustrative purposes and are approximated from the information given by the authors.}}
    \label{fig:positioning}
    \vspace{-0.3in}
\end{figure}

LLMs are a natural candidate to move beyond this scope, given their adaptability across diverse medical tasks and report types \citep{nori2023can,dada-etal-2025-biomedical,wang2026llm_mri_diagnosis}. Yet, studies on general-purpose radiology evaluators remain limited in scope. Existing metrics are largely proposed and validated in isolation, leaving unclear which closed- or open-source models and prompting strategies yield the most reliable judges. We address this with a systematic study of LLM-as-a-judge metrics across open- and closed-source models on two expert-annotated datasets, RadEval and RaTE-Eval, spanning multiple modalities and anatomies (Figure~\ref{fig:positioning}). Furthermore, we introduce VERT, a simple direct-scoring LLM-based metric.

When training data is available, we show that ensembling and fine-tuning yield complementary in-domain gains in correlation with radiologist judgments. Thus, lightweight adaptation remains an important way to achieve efficiency, although prompt-based evaluation using generalist LLMs should also be prioritized.

Finally, we carry out a systematic error-injection and error-breakdown study to analyze, at a granular level, where LLM-based metrics align with and diverge from expert annotations.

Our contributions are as follows:
\begin{itemize}
    \setlength{\topsep}{0pt}
    \setlength{\itemsep}{0pt}
    \item We present a \textbf{systematic study of LLMs as radiology-report judges} across two expert-annotated datasets spanning \textbf{multiple modalities and anatomies}, benchmarking closed- and open-source models and analyzing the impact of prompting strategies from which we distill \textbf{practical recommendations} for practitioners.
    \item We show that \textbf{two recent fine-grained LLM-based methods underperform GREEN} on both benchmarks, while \textbf{VERT}, a simple direct-scoring LLM-based metric, achieves the \textbf{highest correlation with expert judgments}, across the evaluated LLMs and among four prompt variants on the held-out evaluation subsets of both RadEval and RaTE-Eval.
    \item We show that \textbf{ensembling} and \textbf{fine-tuning} provide complementary gains beyond prompting---with latency reduction for the latter---when data is available.
    \item We conduct \textbf{error-injection and error-breakdown analyses} to identify the strengths and failure modes of LLM-based metrics.
    \item We release our code repository \textbf{RadEvalLLM}\footnote{Release upon acceptance.} for reproducibility across the main LLM judges and configurations.
\end{itemize}

\section{Related Work}

\subsection{Traditional and Hybrid Metrics for Radiology}

Early methods rely on lexical and embedding-based metrics such as BLEU \citep{papineni2002bleu}, ROUGE \citep{lin2004rouge} and BERTScore \citep{zhang2020bertscore}, which measure overlap between generated and reference reports without explicitly capturing clinical relevance. To address this limitation, methods such as CheXbert \citep{smit2020chexbert}, and RadGraph \citep{jain2021radgraph} evaluate agreement over predefined labels or structured entity–relation representations. Although better aligned with diagnostic relevance, these approaches depend on external extractors and remain constrained by predefined ontologies. Composite and learned metrics, such as RadCliQ \citep{yu2023radcliq}, combine multiple signals to better approximate expert judgment. However, they rely on labeled data and may suffer from reduced interpretability.

\subsection{LLM-based Metrics for Radiology}

Recent approaches leverage LLMs as evaluators \citep{zheng2023judging,liu2023geval,gu2024survey}. Prior works such as RadFact~\citep{Bannur2024MAIRA2GR}, GREEN~\citep{ostmeier2024green}, FineRadScore~\citep{huang2024fineradscore}, CLEAR~\citep{jiang2025clear}, and CRIMSON~\citep{baharoon2026crimson} use LLMs to assess the clinical factuality of a radiology report against a reference report using different schemas. These approaches demonstrate stronger alignment with expert judgments by capturing nuanced clinical errors. Despite these advances, existing metrics are developed for CXR with limited evaluation across modalities and anatomical regions. Consequently, key questions regarding model selection, prompting strategies, and generalization remain underexplored.






\subsection{Fine-Tuning Models as Evaluators in Radiology}

Several radiology metrics are based on small models finetuned on expert-annotated labels~\citep{smit2020chexbert,jain2021radgraph} or distillation via LLM annotations~\citep{ostmeier2024green,delbrouck2025srrbert,cho2025crepe,baharoon2026crimson}. While most were evaluated on the ReXVal dataset~\citep{PhysioNet-rexval-dataset-1.0.0}, based on CXR, recent results by \citet{cho2025crepe} on the RaTE-Eval dataset \citep{zhao2024ratescore}, covering a diversity of modalities and anatomies, point to a potential vulnerability of these metrics when evaluated outside the scope of CXR, except for GREEN.  

\subsection{LLM-based Error Injection}

Recent studies support LLM-based error injection as a promising way to construct controlled evaluation datasets. \citet{luhtaru2024err} and \citet{latouche2024zero} demonstrate the utility of LLM-generated grammatical errors, while \citet{rao2024rexerr} extend this approach to clinically meaningful errors in radiology reports, using domain expertise to support plausibility. Together, these studies motivate using top-tier LLMs to inject diverse, targeted errors into correct inputs to stress test LLM judges, enabling systematic evaluation beyond naturally occurring errors.

\section{Datasets}
\label{sec:datasets}

In this study, we use RadEval and RaTE-Eval, two recent datasets featuring a higher number of radiology reports and several modalities and anatomies in comparison to the widely used ReXVal~\citep{PhysioNet-rexval-dataset-1.0.0}. Neither dataset provides a dedicated development split. For the prompt-selection experiment, we partition each test set into two equal subsets, using one for prompt selection and the other for evaluation. All other benchmarking experiments use the original test sets to preserve comparability with prior work.

\textbf{RadEval}~\citep{xu2025radeval} consists of 148 chest x-rays. Each study has a ground-truth report and 3 annotated candidate reports. The expert annotations detail the number of significant and insignificant errors found in the candidate report for each of the six error categories from \cite{PhysioNet-rexval-dataset-1.0.0}: \textbf{(a)}~false prediction of a finding, (\textbf{b)}~omission of a finding, \textbf{(c)}~incorrect anatomical location or position, \textbf{(d)}~incorrect severity assessment, \textbf{(e)}~mention of a comparison absent from the reference, \textbf{(f)}~omission of a change noted in a prior study, and a seventh category to capture inarticulate report/grammar. However, they do not contain an overall score for the candidate report. We show the histogram of significant errors in Figure \ref{fig:radeval_dist} of Appendix \ref{sec:appendix_dist}.

\textbf{RaTE-Eval}~\citep{zhao2024ratescore} consists of sentence- and paragraph-level benchmarks spanning 9 imaging modalities and 22 anatomies. In this work, we use the paragraph-level benchmark containing a total of 1856 reference reports (i.e., 444 as testset) paired with a candidate report and an expert-annotated score ranging from 0 to 5. We provide the histogram of normalized scores (0--1) in Figure \ref{fig:rateeval_dist} of Appendix \ref{sec:appendix_dist}.

\section{Methodology}

\subsection{Correlation With Human Judgments}
\subsubsection{Zero-shot Prompting}
\label{sec:prompt_design_methods}

We conduct a series of controlled experiments to identify the zero-shot prompt and model combinations that are most effective at radiology report evaluation. In addition to evaluating existing prompting strategies, we systematically compare multiple score aggregations to assess how prompt design influences alignment with expert judgments and to examine the effect of direct score prediction versus derived or structured scoring schemes.

Specifically, we evaluate four zero-shot prompt variants using GPT-4.1-mini and Gemini-2.5-flash: (1) \textbf{GREEN}, a chain-of-thought prompt that identifies clinically significant and insignificant errors and computes a score post hoc; (2) \textbf{VERT}, our proposed metric, which instead directly prompts the model to produce an overall accuracy score along GREEN's error-count outputs; (3) a \textbf{formula}-based prompt, which instructs the model to compute a score using a predefined weighted equation over matched findings and error types; and (4) a \textbf{rubric}-based prompt, which replaces the formula with a discrete scoring rubric. We focus on these prompts to test whether model-generated scores align with human ratings, and whether providing the model with a scoring formula or rubric can improve that alignment further. Since we are selecting the best prompt method in this experiment specifically, we split the test sets of RadEval and RaTE-Eval 50/50 for development and held-out test purposes. Full prompts are available in Appendix~\ref{sec:appendix_prompts}. 


We then further evaluate our proposed zero-shot prompt, VERT, across multiple large language models, including both closed- and open-source variants, to assess robustness across model families. 
We measure the correlation with human annotations on the full testsets of both RadEval and RaTE-Eval. We also compare it against established baselines: RadFact~\citep{Bannur2024MAIRA2GR}, GREEN~\citep{ostmeier2024green}, GREEN-EC (sum of error counts), FineRadScore~\citep{huang2024fineradscore}, and CRIMSON~\citep{baharoon2026crimson}. We use the implementations from their GitHub repositories.

We excluded lexical metrics from our baselines because previous works~\cite{xu2025radeval,ostmeier2024green,huang2024fineradscore} have established weaker correlations with expert judgment.

We focus on the absolute value\footnote{Correlation of a metric between 0-1 with the sum of errors yields a negative value while it is positive when using a holistic expert score. This sign difference is irrelevant here.} of Kendall's $|\tau|$ to ease comparison of our results against prior work, even if it tends to lead to noticeably lower scores than Pearson's $r$ or Spearman's $\rho$ correlations\footnote{For example, a strong-positive correlation of $r\approx\rho\approx0.7$ can correspond to $\tau\approx0.5$.}.

\subsubsection{Ensembling}
We leverage two ensembling methods: \textbf{averaging} of scores and \textbf{linear regression}. For the former, we simply average metric scores while, for the latter, we fit a linear regression on LLM scores over 500 samples from the RaTE-Eval training set. We used three frontier models for our ensembles: GPT-4.1, Gemini 2.5 Flash (low), and Claude Opus 4.6 (low).

\subsubsection{Few-shot Prompting}
We investigate whether the inclusion of few-shot examples and error descriptions improves correlation with human annotations. We evaluate 3-, 5-, 10- shots sampled at random from either the RadEval dataset or the RaTE-Eval trainset across 5 runs (i.e., seeds of 42 to 46). Each shot includes the candidate report, the reference report, and an overall score based on human assessments. In addition to the random few-shot prompts, we test four curated prompts: (1) \textbf{rad-err} includes six human annotations from RadEval, one per error type. (2) \textbf{rate-err} includes four examples from RaTE-Eval's trainset. Since human assessments for RaTE-Eval do not include error counts (only overall scores from 0-5), we inject reference reports with errors of type (a) and (b) using GPT-4.1. (3) \textbf{rad-err-10 human} features error descriptions for each error type and 10 random RadEval. (4) \textbf{rate-err-10 VERT} features error descriptions for each error type and 10 random RaTE-Eval shots with assessments generated using GPT-4.1-mini VERT. Full prompts are available in Appendix~\ref{sec:appendix_prompts}.

\subsubsection{LoRA Fine-Tuning}
We finetune \texttt{Qwen3-30B-A3B-Instruct-2507} \citep{yang2025qwen3}---a mixture-of-expert (MoE) architecture with 8 activated experts out of 128---leveraging the QLoRA low-rank adaptation method \citep{dettmers2023qlora}. The base model is frozen and quantized in 4 bits in memory (with \textit{bfloat16} for computing). Then, trainable LoRA adapters \citep{hu2022lora} of rank 32 ($\alpha=64$) are injected into each linear layer, except the MoE router layers. The optimizer is AdamW~\citep{loshchilov2019adamw} including a cosine scheduler with a peak learning rate of $2 \times 10^{-4}$ trained for 5 epochs fixed \textit{a priori} based on available budget and best practices. We used the user-turn template with the first two lines from the VERT prompt followed by the reference and candidate reports, and then we applied the assistant-turn template to wrap the normalized score between 0.0 and 1.0 with only 1 decimal. A score histogram is provided in Figure \ref{fig:rateeval_dist} in Appendix \ref{sec:appendix_dist}. Applying the chat template enables full compatibility with vLLM~\citep{kwon2025vllm}.

We finetune 1,486 report paragraphs from the RaTE-Eval training set, which we split 90:10 for training and validation purposes, respectively.

\subsection{Error Injection and Type Breakdown}

In addition to the correlation analysis, we study the performance of model and prompt combinations at correctly detecting and categorizing errors. Specifically, we compare the number and type of clinically significant errors identified by the system against those identified by human annotators. While error counts for each category are included in human assessments in RadEval, expert assessments in RaTE-Eval only provide an overall quality score. Therefore, in order to study error detection and categorization in RaTE-Eval, we inject clinically significant errors into reference radiology reports using GPT-4.1. We consider two error types: \textbf{false prediction} (category~a), and \textbf{missing finding} (category~b). We focus on these error types as they are the most frequent in RadEval (Figure \ref{fig:radeval_error_type_tau}). For each report, the model is instructed to introduce exactly $k \in \{1, 2, 3\}$ clinically significant errors of the target type while preserving all other content. After injection, each perturbed report is passed through a separate validation step in which GPT-4.1 assesses whether the injected change is anatomically and clinically plausible given the imaging modality and body region. Modifications that fail validation are regenerated up to 5 times. Reports that remain implausible after all attempts are excluded from the experiment. An expert reading of a 100 perturbed cases found 6 that were implausible. Full prompts can be found in Appendix \ref{sec:appendix_injection_prompts}.

\begin{table*}[t]
\centering
\small
\setlength{\tabcolsep}{0.6em}
\renewcommand{\arraystretch}{1.2}
\begin{tabular}{lcccccc}
\toprule
Model & RadFact & GREEN & GREEN-EC & FineRadScore & CRIMSON & VERT \\
\midrule
\multicolumn{7}{c}{\textbf{RadEval}\ \ ---\ \ Error Annotations} \\
\midrule
GPT-4.1 mini & 0.295 & 0.332 & \textbf{0.561} & 0.173 & 0.264 & \underline{0.371} \\
Gemini 2.5 Flash (low) & - & 0.319 & \textbf{0.507} & \underline{0.378} & 0.263 & 0.338 \\
Claude Sonnet 4.6 (low) & - & 0.293 & \textbf{0.510} & 0.233 & 0.201 & \underline{0.329} \\
\midrule
\multicolumn{7}{c}{\textbf{RaTE-Eval}\ \ ---\ \ Expert Scores (0--5)} \\
\midrule
GPT-4.1 mini & 0.312 & \underline{0.420} & 0.197 & 0.137 & 0.218 & \textbf{0.447} \\
Gemini 2.5 Flash (low) & - & \underline{0.344} & 0.186 & 0.212 & 0.269 & \textbf{0.427} \\
Claude Sonnet 4.6 (low) & - & \underline{0.432} & 0.246 & 0.195 & 0.295 & \textbf{0.453} \\
\bottomrule
\end{tabular}
\caption{Kendall's $|\tau|$ across prompts and models on all test sets. RadFact's Python library is only compatible with OpenAI. \textbf{Bold} and \underline{underlined} indicate the best and second best result per row respectively.}
\label{tab:lit_metric_comparison}
\vspace{-0.1in}
\end{table*}

\section{Results}

\subsection{Correlation With Human Judgments}
\label{sec:zero_prompt_results}

\subsubsection{Model-generated scores correlate more with expert judgments}


\begin{table}[h!]
\small
\setlength{\tabcolsep}{0.45em}
\renewcommand{\arraystretch}{1.2}
\centering
\begin{tabular}{lcccc}
\toprule & \multicolumn{2}{c}{GPT-4.1-mini} & \multicolumn{2}{c}{Gemini 2.5 Flash} \\
\cmidrule(lr){2-3}
\cmidrule(lr){4-5} Metric & RadEval & Rate-Eval & RadEval & Rate-Eval \\
\midrule
GREEN & 0.350 & 0.409 & 0.346 & 0.313 \\
Formula & 0.081 & 0.375 & 0.210 & 0.335 \\
Rubric & 0.292 & \textbf{0.445} & 0.286 & 0.119 \\
VERT & \textbf{0.369} & 0.443 & \textbf{0.357} & \textbf{0.386} \\
\bottomrule
\end{tabular}
\caption{ Comparison of GREEN, Formula, Rubric, and VERT across $10k$ paired-bootstrap samples on the 50\% development set. Standard deviations range from $0.045$ to $0.070$. \textbf{Bold} indicates the highest mean for each model--dataset pair. }
\label{tab:green_variants_vert_comparison}
\end{table}

We first compare the performance of zero-shot prompts with different score assignment instructions. Our proposed prompt, VERT, achieves the strongest correlation with the annotations of human experts for both RadEval and Rate-Eval compared to GREEN---as well as the formula- and rubric-based prompts (Tables \ref{tab:green_variants_vert_comparison} and \ref{tab:green_vert_prob}). The results also suggest that including a formula or a rubric might decrease performance in most cases, except for \textit{Rubric} on Rate-Eval with GPT-4.1-mini.

\begin{table}[h]
\small 
\setlength{\tabcolsep}{0.45em}
\renewcommand{\arraystretch}{1.2}
\centering
\begin{tabular}{lcccc}
\toprule
& \multicolumn{2}{c}{GPT-4.1-mini} & \multicolumn{2}{c}{Gemini 2.5 Flash} \\
\cmidrule(lr){2-3} \cmidrule(lr){4-5} Metric & RadEval & Rate-Eval & RadEval & Rate-Eval \\
\midrule
GREEN & 0.323 & 0.407 & 0.297 & 0.366 \\
VERT & \textbf{0.358} & \textbf{0.494} & \textbf{0.329} & \textbf{0.461} \\
$P(\mathrm{V}>\mathrm{G})$ & 0.961 & 0.998 & 0.811 & 0.992 \\
\bottomrule
\end{tabular}
\caption{Comparison of GREEN and VERT scores across $10k$ paired-bootstrap samples on the 50\% held-out test set. \textbf{Bold} indicates the higher score for each model--dataset pair. $P(\mathrm{V} > \mathrm{G})$ denotes the probability that \textbf{V}ERT outperforms \textbf{G}REEN. } \label{tab:green_vert_prob}
\end{table}

We further validated the robustness of VERT to prompt sensitivity in Appendix \ref{app:prompt_variants}, and provided both Spearman and Pearson correlations in Appendix \ref{app:pearson_spearman}.

\subsubsection{Different zero-shot judges for different datasets}

We compare our zero-shot VERT prompt against four metrics from prior work (RadFact, GREEN, FineRadScore and CRIMSON) on both RadEval and RaTE-Eval  (Table \ref{tab:lit_metric_comparison}). We also include GREEN-EC, the sum of the error counts produced by the model. We prompted GPT-4.1 mini, Gemini 2.5 Flash, and Claude Sonnet 4.6.


Despite being top performers on ReXVal, FineRadScore and CRIMSON are underperforming across models on both datasets in our setting. Finally, we observe that \textit{VERT is overall the most consistent bounded metric} on both datasets, while the unbounded GREEN-EC is specifically better on RadEval but worst on RaTE-Eval.

\subsubsection{Best LLMs as Radiology Evaluators}


\begin{table}[h]
\small
\setlength{\tabcolsep}{0.4em} \renewcommand{\arraystretch}{1.2}
\centering
\begin{tabular}{lccc}
\toprule
\textbf{Models} & \makecell{\textbf{GREEN}} & \makecell{\textbf{GREEN-EC}} & \makecell{\textbf{VERT}} \\
\midrule
GPT-4.1 mini & 0.420 & 0.197 & 0.447 \\
GPT-4.1 & 0.365 & 0.222 & 0.452 \\
GPT-5 & 0.333 & 0.138 & 0.409 \\
GPT-5 mini & 0.350 & 0.221 & 0.330 \\
GPT-5.6 Luna & 0.347 & 0.197 & 0.365 \\
GPT-5.6 Sol & 0.427 & \textbf{0.296} & 0.447 \\
Gemini 2.5 Flash & 0.344 & 0.186 & 0.427  \\
Opus 4.6 & \textbf{0.449} & 0.241 & \textbf{0.463} \\
Sonnet 4.6 & \underline{0.432} & \underline{0.246} & \underline{0.453} \\
Qwen3 30B & 0.285 & 0.059$^\dagger$ & 0.209 \\
\bottomrule
\end{tabular}
\caption{Kendall's $|\tau|$ on all RaTE-Eval's test set across 10 LLMs. \textbf{Bold} and \underline{underline} are the best and second-best value, respectively, per metric. All reasoning LLMs are set to \textit{low}. $\dagger$ denotes a \textit{p}-value above $0.01$.} \label{tab:model_green_vert_results}
\vspace{-0.1in}
\end{table}

In Table \ref{tab:model_green_vert_results}, we prompt ten models with GREEN and VERT. Generally, we notice that Claude models are the best performing, followed by the GPT-5.6 and GPT-4.1 families.

\subsubsection{Thinking does not improve correlation with expert judgments}

\begin{table}[h]
\small
\setlength{\tabcolsep}{0.5em}
\renewcommand{\arraystretch}{1.2}
\centering
\begin{tabular}{lcccc}
\toprule
\textbf{Model} &
\makecell{\textbf{None}} &
\makecell{\textbf{Low}} &
\makecell{\textbf{Medium}} &
\makecell{\textbf{High}} \\
\midrule
GPT-5 &
-- &
0.333 &
\textbf{0.367} &
0.351 \\

GPT-5 mini &
-- &
\textbf{0.350} &
0.348 &
0.275 \\

GPT-5.6 Luna &
-- &
0.347 &
0.341 &
\textbf{0.373} \\

GPT-5.6 Sol &
-- &
\textbf{0.427} &
0.369 &
0.395 \\

Claude Opus &
0.441 &
\textbf{0.449} &
0.359 &
0.200 \\

Claude Sonnet &
\textbf{0.434} &
0.432 &
0.299 &
0.205 \\

Qwen3 30B &
0.041$^\dagger$ &
-- &
\textbf{0.285} &
-- \\

Qwen3 4B &
0.043$^\dagger$ &
-- &
\textbf{0.048}$^\dagger$ &
-- \\
\bottomrule
\end{tabular}
\caption{Kendall's $|\tau|$ correlation of GREEN on all RaTE-Eval's test set across thinking levels. The \textit{medium} reasoning level is the \textit{thinking} variant for Qwen3 models. \small{Dashes indicate unavailable configurations. \textbf{Bold} denotes the best value per model. The dagger $\dagger$ denotes a \textit{p}-value above $0.01$.}}
\label{tab:thinking_levels_all_metrics}
\vspace{-0.1in}
\end{table}

We display nine open- and closed-source thinking models prompted with the GREEN metric approach in Table \ref{tab:thinking_levels_all_metrics} covering all reasoning efforts (none, low, medium and high).

Overall, we notice mixed results from extending the number of reasoning tokens. Only Qwen3 30B benefits from longer reasoning traces, while still performing lower than the others on GREEN. We observe that the smaller open-source model---i.e., Qwen3 4B---does not correlate with expert scores in any setting. For both GPT and Claude models, correlation tends to decrease in the \textit{high} thinking setting.

\subsubsection{Regression-based ensemblings improve correlation with expert judgments}

We compute several ensembles of GPT-4.1, Claude Opus 4.6 (low) and Gemini 2.5 Flash (low) in Table~\ref{tab:ensemble_results}. We select the best correlation scores for each metric among those three models as baselines (Table \ref{tab:model_green_vert_results}). Two ensembling methods were performed: average of scores and linear regression. The latter is fitted using 500 samples from the RaTE-Eval trainset. All ensembles are performed individually per score, except for the linear regression \textit{All} which is done on all available features from GREEN or VERT.

\begin{table}[h!]
\setlength{\aboverulesep}{0.1ex}
\setlength{\belowrulesep}{0.1ex}
\small
\setlength{\tabcolsep}{0.4em}
\renewcommand{\arraystretch}{1.0}
\centering
\begin{tabular}{l|l|ccc}
\toprule
& \textbf{Metric} & \textbf{Baselines} & \textbf{Average} & \makecell{\textbf{Linear}\\\textbf{Regression}} \\
\midrule
\multirow{4}{*}{\textbf{GREEN}} & SCORE & 0.449 & 0.421 & \textbf{0.452} \\
& EC    & 0.241 & 0.256 & \textbf{0.265} \\
 & All   & -- & -- & \textbf{0.486} \\
\midrule
\multirow{2}{*}{\textbf{VERT}}  & SCORE & 0.463 & 0.480 & \textbf{0.489} \\
& All   & --     & --     & \textbf{0.509} \\
\bottomrule
\end{tabular}
\caption{Kendall's $|\tau|$ between ensemble-aggregated GREEN/VERT metrics on RaTE-Eval's full test set. \small{Each row uses only the corresponding metric scores from each model, except for \textit{All} which leverages all scores from GREEN or VERT---i.e., matched findings, error counts, and scores. \textbf{Bold} indicates the best result per metric.}}
\label{tab:ensemble_results}
\vspace{-0.1in}
\end{table}

Averaging improves performance against baseline only in the case of GREEN EC and VERT, while linear regression improves all scores. Combining all available features yields the highest correlations: Kendall’s $|\tau|$ reaches 0.486 for GREEN and 0.509 for VERT, compared with their strongest individual score baselines.

\subsubsection{Few-shot Prompting}

\begin{table*}[ht] \centering \small \renewcommand{\arraystretch}{1.1} \setlength{\tabcolsep}{5pt} \setlength{\aboverulesep}{0.5ex} \setlength{\belowrulesep}{0.5ex} \begin{tabular}{c|c|ccccc|ccccc} \toprule \multicolumn{1}{c|}{} & \multicolumn{1}{c|}{} & \multicolumn{5}{c|}{\underline{RadEval Shots}} & \multicolumn{5}{c}{\underline{RaTE-Eval Shots}} \\ & \textbf{0-shot} & \textbf{3} & \textbf{5} & \textbf{10} & \textbf{rad-err} & \makecell{\textbf{rad-err} \\ \textbf{10 human}} & \textbf{3} & \textbf{5} & \textbf{10} & \textbf{rate-err} & \makecell{\textbf{rate-err} \\ \textbf{10 VERT}} \\ \midrule \textbf{RadEval} & 0.359 & \makecell{0.373 \\ {\color{green!55!black}\scriptsize +1.4\%}} & \makecell{0.370 \\ {\color{green!55!black}\scriptsize +1.1\%}} & \makecell{0.405 \\ {\color{green!55!black}\scriptsize +4.6\%}} & \makecell{0.393 \\ {\color{green!55!black}\scriptsize +3.4\%}} & \makecell{\textbf{0.445} \\ {\color{green!55!black}\scriptsize +8.6\%}} & \makecell{0.341 \\ {\color{red!70!black}\scriptsize -1.8\%}} & \makecell{0.345 \\ {\color{red!70!black}\scriptsize -1.4\%}} & \makecell{0.349 \\ {\color{red!70!black}\scriptsize -1.0\%}} & \makecell{\textbf{0.444} \\ {\color{green!55!black}\scriptsize +8.5\%}} & \makecell{0.372 \\ {\color{green!55!black}\scriptsize +1.3\%}} \\ \midrule \textbf{RaTE-Eval} & 0.445 & \makecell{0.479 \\ {\color{green!55!black}\scriptsize +3.4\%}} & \makecell{0.480 \\ {\color{green!55!black}\scriptsize +3.5\%}} & \makecell{\textbf{0.490} \\ {\color{green!55!black}\scriptsize +4.5\%}} & \makecell{0.462 \\ {\color{green!55!black}\scriptsize +1.7\%}} & \makecell{0.453 \\ {\color{green!55!black}\scriptsize +0.8\%}} & \makecell{0.485 \\ {\color{green!55!black}\scriptsize +4.0\%}} & \makecell{0.483 \\ {\color{green!55!black}\scriptsize +3.8\%}} & \makecell{\textbf{0.487} \\ {\color{green!55!black}\scriptsize +4.2\%}} & \makecell{0.410 \\ {\color{red!70!black}\scriptsize -3.5\%}} & \makecell{0.460 \\ {\color{green!55!black}\scriptsize +1.5\%}} \\ \bottomrule \end{tabular} \caption{ Mean Kendall's $|\tau|$ between the VERT output score and human judgments for GPT-4.1-mini. The maximum standard deviation across all five runs and configurations is $0.011$. \textbf{Bold} indicates the best mean per row and type of example, while colored values show the absolute percentage-point difference from the zero-shot baseline. } \label{tab:prompt_comparison_combined_v2} \vspace{-0.2in} \end{table*}

We investigate the performance of adding random 3-, 5-, and 10-shots, error descriptions, and error-specific curated shots on both RadEval and RaTE-Eval (Table~\ref{tab:prompt_comparison_combined_v2}). We choose GPT-4.1-mini as our base model, since it is affordable while still achieving satisfactory performance (Tables~\ref{tab:lit_metric_comparison} and \ref{tab:green_variants_vert_comparison}).

Few-shot improvements depend on the example source and prompt design. On RadEval, curated error-specific prompts achieve the highest correlations, with rad-err-10-human and rate-err reaching 0.445 and 0.444, respectively, compared with 0.359 for zero-shot prompting. On RaTE-Eval, including random shots from RadEval or RaTE-Eval outperforms zero-shot (0.445) and all four curated prompts (ranging 0.410 and 0.462). 10 random RadEval or RaTE-Eval examples perform best (0.490 and 0.487). Random RadEval-sourced examples improve correlation on both datasets, whereas random RaTE-Eval-sourced examples improve correlation only on RaTE-Eval. Increasing the number of examples does not consistently improve performance, and no configuration performs best on both datasets.

\subsubsection{LoRA Fine-Tuning for RaTE-Eval}

\begin{figure}[h]
    \centering
    \includegraphics[width=0.95\linewidth]{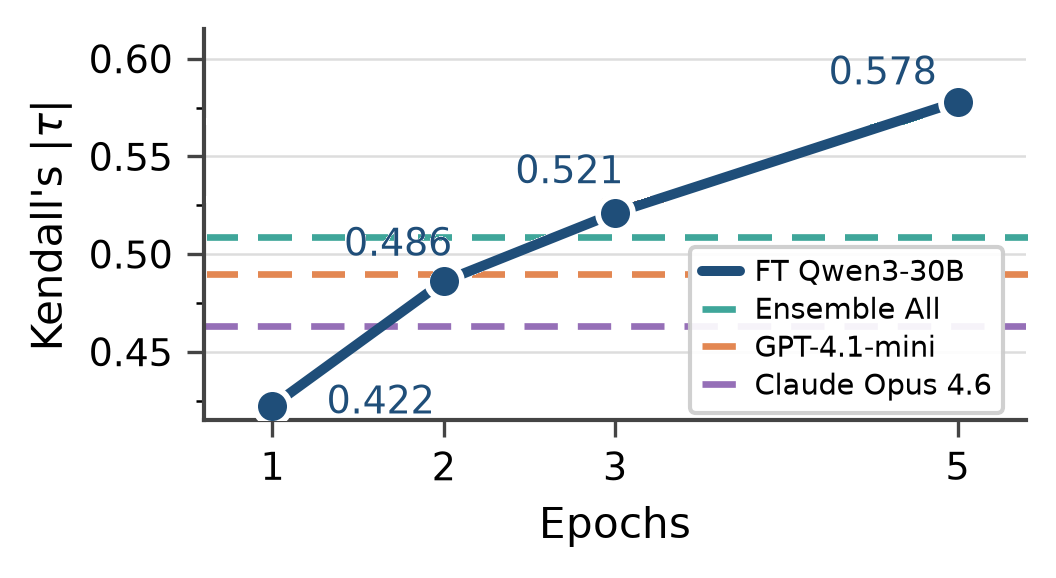}
    \vspace{-0.2in}
    \caption{Correlation on RaTE-Eval testset of finetuned Qwen3 for 5 epochs compared to VERT baselines.}
    \label{fig:ftlora_qwen3_30b}
    \vspace{-0.1in}
\end{figure}
We finetune \texttt{Qwen3-30B-A3B-Instruct-2507} on RaTE-Eval in Figure \ref{fig:ftlora_qwen3_30b}, leading to correlations with expert judgment surpassing all previous LLM-based evaluators at $0.578$. We also evaluate our finetuned model on the RadEval testset scoring a $|\tau|$ of $0.185$, which indicates a weak out-of-domain generalization. We provide training and validation losses in Appendix \ref{app:training_dynamics}.

At inference time, the finetuned model processes RaTE-Eval's testset in 36 seconds on an A100 GPU, while Claude Opus 4.6 (low) using VERT takes about 22.3 minutes with the Anthropic API, a \textbf{$37.2\times$ reduction in computation time}\footnote{Since GPU runtime is unavailable with closed-source models, this number is limited to the API latency. Yet, the order of magnitude remains representative.}.

\subsection{Error Injection and Type Breakdown}

\begin{figure}[h]
\centering
\includegraphics[width=\linewidth]{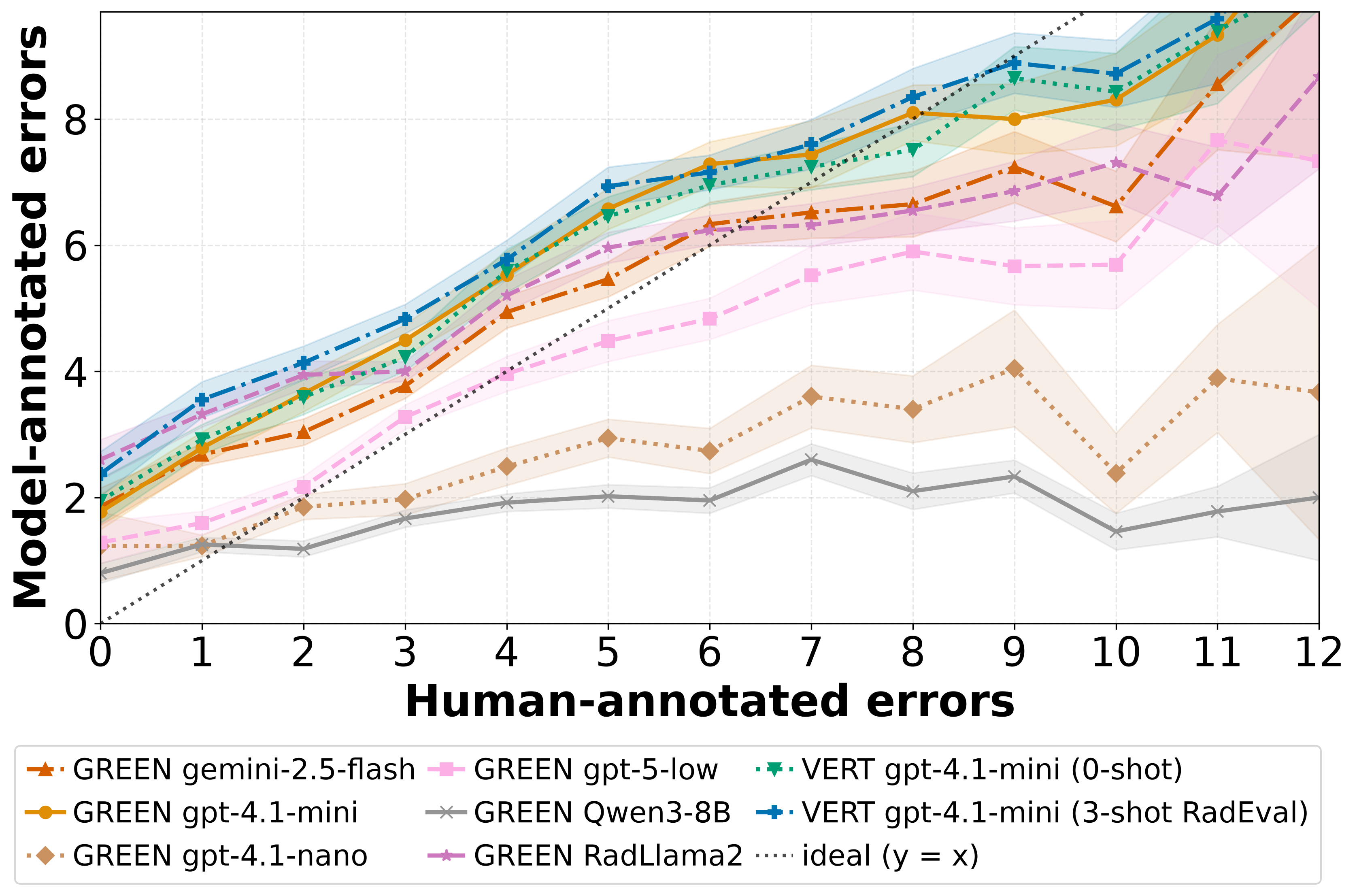}
\vspace{-0.3in}
\caption{Average \# expert-annotated significant errors vs. LLM-detected errors on RadEval.}
\vspace{-0.2in}
\label{fig:green_vs_human_bin_means}
\end{figure}

We study whether zero-shot and few-shot LLM-as-a-judge methods can retrieve the same number of clinically significant errors as expert annotators. For RadEval, we test zero-shot with a variety of models and the few-shot method most correlated with human judgments (VERT 3-shot RadEval). We restrict our analysis to reports containing between 1 and 12 human-annotated errors to ensure sufficient representation within RadEval (Fig.~\ref{fig:radeval_dist}).

\begin{figure}[h]
    \centering
    \includegraphics[width=\linewidth]{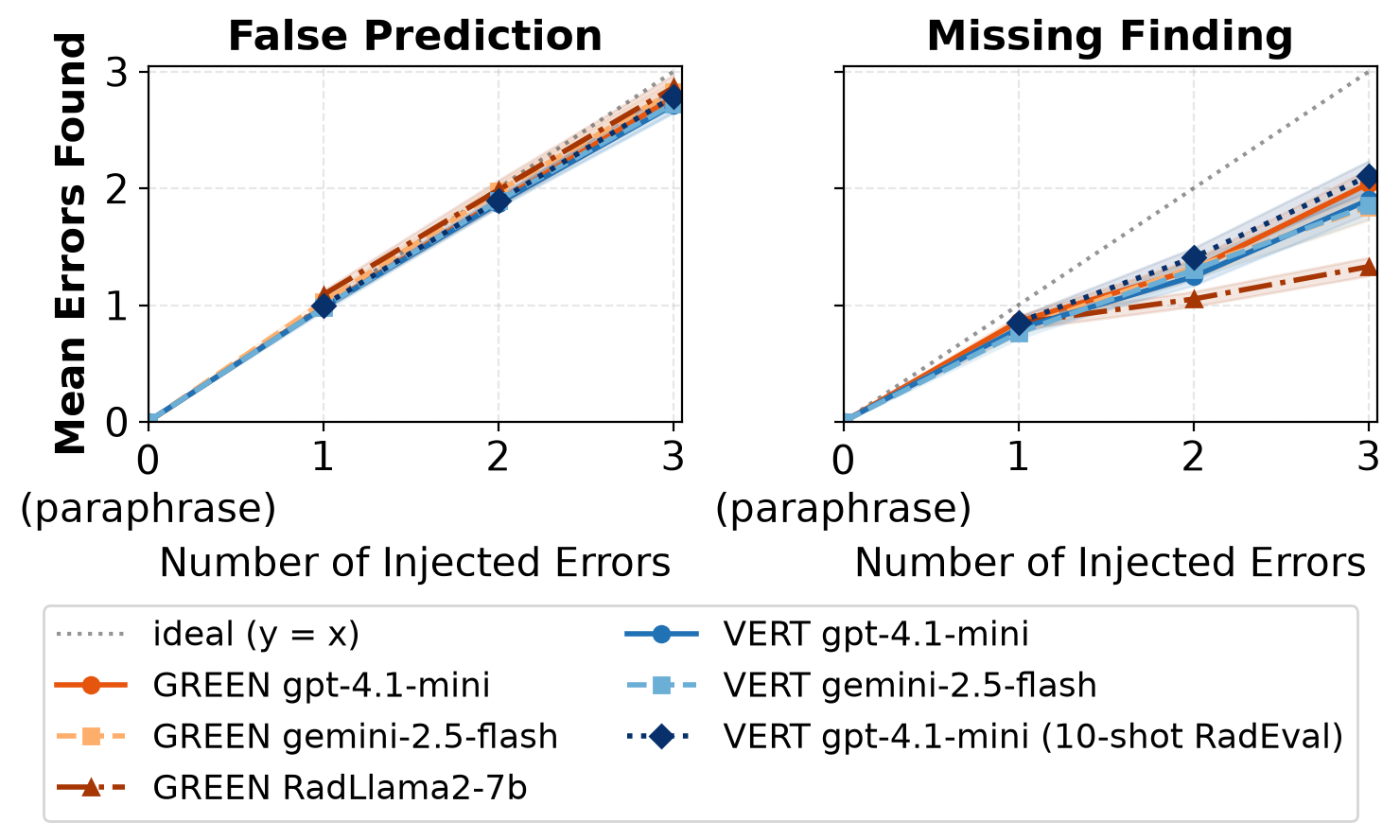}
    \caption{Average \# LLM-detected significant errors vs. \# injected errors on RaTE-Eval.}
    \label{fig:errors_found_vs_injected}
\end{figure}

The majority of model--prompt combinations overestimates the number of errors when there are fewer than 6 human-annotated errors and underestimates when there are more than 6 (Figure ~\ref{fig:green_vs_human_bin_means}). This is true even for recent proprietary models like GPT-5 and Gemini 2.5 Flash. GREEN Qwen3 8b is the worst performing of the group. GPT-4.1-mini-based approaches are the ones that most closely match human annotations, with VERT GPT-4.1-mini as the best performing one. In the case of RaTE-Eval (Figure~\ref{fig:errors_found_vs_injected}), all approaches find most of the injected errors of type (a), false prediction. However, they all underestimate the number of injected errors of type (b), missing finding, when there is more than 1 error of that type. Despite the fact that we inject errors with GPT-4.1, we do not notice a specific bias towards this model at detection time compared to other models.

\begin{figure}[h]
    \centering    \includegraphics[width=0.8\linewidth]{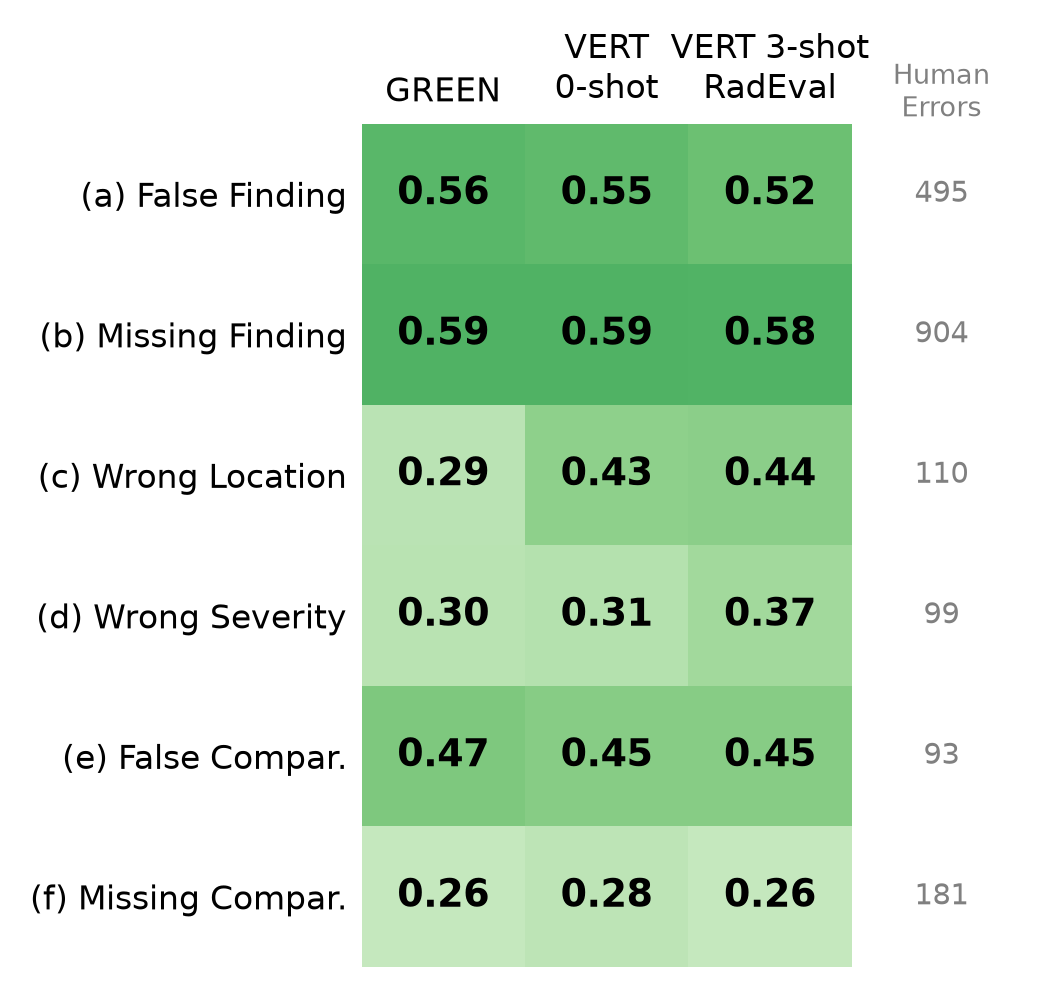}
    \vspace{-0.1in}
    \caption{$|\tau|$ for each error type \textit{a--f} on RadEval.}
    \label{fig:radeval_error_type_tau}
    \vspace{-0.1in}
\end{figure}

Given these results, we take a closer look at the alignment with expert judgment at the error-type level. We report the Kendall's correlation between predicted and expert error counts per type. On RadEval (Figure~\ref{fig:radeval_error_type_tau}), all approaches align well with experts on (a) and (b) ($|\tau|$: 0.52--0.59), which are also the best supported. Agreement is lower on the rarer types (c--f) ($|\tau|$: 0.26--0.47). VERT improves by +48\% over GREEN on (c) with smaller gains on (d) and (f), and few-shot prompting tends to redistribute those gains---e.g., +20\% on (d).

\begin{table}[h]
    \renewcommand{\arraystretch}{1.0}
    \setlength{\tabcolsep}{2pt}
    \centering
    \small
    \begin{tabular}{lcc}
    \toprule
    \textbf{Method} & \textbf{(a) False Finding} & \textbf{(b) Missing Finding} \\
    \midrule
    GREEN & 0.91 & 0.75 \\
    VERT & 0.89 & 0.75 \\
    VERT \tiny{10-shot RadEval} & 0.87 & 0.74 \\
    \midrule
    Injected errors & 2064 & 2130 \\
    \bottomrule
    \end{tabular}
    \caption{$|\tau|$ for injected errors on RaTE-Eval for gpt-4.1-mini.}
    \label{tab:rate_eval_error_type_kendall}
\end{table}

The RaTE-Eval dataset doesn't provide annotations per error types. Nonetheless, we use injected errors from the previous experiment, of which results exhibit a different pattern in Table~\ref{tab:rate_eval_error_type_kendall}. Zero- and few-shot approaches achieve an $|\tau|$ of 0.87--0.91 for error type (a), but only 0.74--0.75 for type (b). The slight decrease in correlation for (a) going from GREEN to VERT with few-shots is similar to the previous trend of (a) on RadEval.

Furthermore, we show a breakdown by imaging modality and anatomy in the Appendix \ref{app:modality_anatomy} showing weaker correlations for MRI and low-frequency anatomy locations, respectively.

\section{Recommendations and Takeaways for Radiology LLM Judges}
\paragraph{Match scoring scheme to annotation target.} No single scoring formulation is universally best (Table~\ref{tab:lit_metric_comparison}). Direct scoring provides a consistent bounded alternative across the two evaluated annotation regimes.
\paragraph{Start simple, limit reasoning effort.} VERT achieves the highest average correlation among the four evaluated prompt variants on both datasets (Tables~\ref{tab:green_variants_vert_comparison}, \ref{tab:green_vert_prob} and \ref{tab:variants}). Higher reasoning effort does not consistently improve correlation under GREEN-style prompting (Table~\ref{tab:thinking_levels_all_metrics}).
\paragraph{Benchmark models rather than relying on recency.} Opus 4.6 and Sonnet 4.6 achieve the highest average correlations on RaTE-Eval among the evaluated models, followed by GPT-5.6 Sol and GPT-4.1 (Table~\ref{tab:model_green_vert_results}).
\paragraph{Validate example source and curation.} Few-shot gains depend on example source and prompt design, with asymmetric cross-dataset transfer. More examples do not consistently improve correlation; select configurations on development data rather than assuming that in-domain examples or larger shot counts are preferable (Table~\ref{tab:prompt_comparison_combined_v2}).
\paragraph{Fine-tune for specialization, validate transfer.} LoRA on Qwen3-30B yields the strongest in-domain correlation, but transfers poorly to RadEval ($|\tau|=0.185$). Its observed inference latency is 37.2$\times$ lower than the proprietary baseline under our evaluated deployment setups.

\section{Conclusion}

A central finding of this work is that two recent fine-grained radiology evaluation methods do not improve alignment with radiologist judgment over GREEN in our multi-modality benchmarks. Instead, our systematic study identifies direct scoring (VERT) as a simple, effective alternative, achieving the highest correlation averaged across evaluated LLMs among four prompt variants on both RadEval and RaTE-Eval. These findings highlight that greater evaluation complexity does not necessarily yield better alignment with expert judgment. Ensembling and fine-tuning offer further in-domain gains when annotated data is available, although the fine-tuned model transfers poorly across datasets.

Finally, our synthetic error analysis shows that correlation with expert judgment does not guarantee accurate error counting or categorization. Evaluated judges misestimate error counts and exhibit weaker agreement on less frequent error categories. Synthetic perturbation experiments reveal particular difficulty counting multiple omissions. These findings motivate benchmarks that assess error detection and categorization alongside overall report quality.


\section*{Limitations}

Our findings should be interpreted in the context of several limitations. First, although our evaluation spans two benchmarks covering multiple imaging modalities and anatomies, it is constrained by the publicly available datasets available for radiology report evaluation. As additional benchmarks become available, it will be valuable to verify whether the trends observed in this work continue to hold across clinical settings and reporting practices.

Second, our study focuses on reference-based radiology report evaluation. While this setting is the standard protocol, future work could investigate how LLM judges behave in complementary settings, such as image-grounded evaluation or assessment without reference reports.

Third, we evaluate a diverse collection of open- and closed-source LLMs under a unified experimental protocol. Given the rapid pace of foundation model development, individual rankings may evolve as newer models become available. Nevertheless, we expect the broader observations regarding prompting strategies, scoring formulations, adaptation methods, and evaluation protocols to remain applicable.

Fourth, perturbations are generated and validated by the same model family (GPT-4.1). However, this limitation is mitigated by the results of Figure~\ref{fig:green_vs_human_bin_means} showing no clear advantage of this LLM against the others. Only fabrication (a) and omission (b) error types were injected.

Finally, our analyses show that current LLM judges correlate well with expert assessments at the report level while remaining less reliable at fine-grained error attribution, particularly for omissions and less frequent error categories. We view this as a limitation of the current evaluation ecosystem rather than of any individual metric, highlighting the need for richer expert benchmarks that jointly evaluate overall report quality and detailed clinical-error localization.

\section*{Ethical Considerations} Automatic evaluation metrics for radiology reports are research tools intended to support method development, not clinical decision-making, and they should not replace radiologist review or be used to certify report-generation systems for clinical deployment. As with any automatic metric, results should be validated on the target clinical setting before use. Our error-injection procedure produces synthetic, intentionally altered reports for controlled analysis only; they are derived from public, de-identified benchmarks and are not intended for clinical use or dissemination. We also note that applying proprietary API-based judges to real clinical reports raises data-governance considerations, which motivates our study of locally deployable open-weight alternatives. All datasets used are publicly available, de-identified, and used consistently with their intended research purpose.

\bibliography{custom}

\appendix

\section{Dataset Statistics}
\label{sec:appendix_dist}

\begin{figure}[h]
\centering
\includegraphics[width=0.6\linewidth]{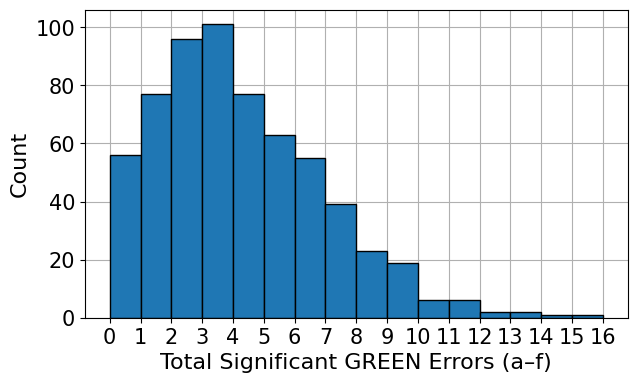}
\caption{Histogram of error annotations in RadEval.}
\label{fig:radeval_dist}
\end{figure}

\begin{figure}[h]
\centering
\includegraphics[width=0.6\linewidth]{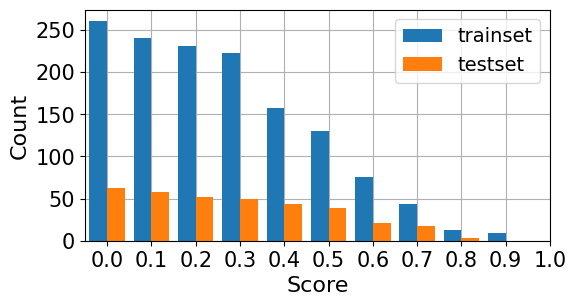}
\caption{Histogram of normalized expert-annotated scores in RaTE-Eval.}
\label{fig:rateeval_dist}
\end{figure}

See Figures ~\ref{fig:radeval_dist} and ~\ref{fig:rateeval_dist}.

\newpage
\section{Complementary Results}

\subsection{VERT Prompt Perturbation Experiment}
\label{app:prompt_variants}

\begin{table}[h]
\centering
\begin{tabular}{lcc}
\toprule
\textbf{Variant} & \boldmath$|\tau|$ & \textbf{std} \\
\midrule
VERT & 0.452 & 0.030 \\
\ \ - Varying wording & 0.464 & 0.031 \\
\ \ - Instruction order   & 0.459 & 0.031 \\
\ \ - Condensing instructions   & 0.452 & 0.031 \\
\bottomrule
\end{tabular}
\caption{Performance of three prompt variants of VERT on RaTE-Eval indicating robustness to prompt sensitivity, i.e., all scores are within the bootstrap \textit{std}.}
\label{tab:variants}
\end{table}

We ran an ablation using GPT-4.1 on prompt sensitivity by making three variants of VERT's prompt in Table~\ref{tab:variants}. Results indicate a small variation (<0.012) overall, in line with recent work that demonstrated a higher robustness to prompt variations with modern LLMs \citep{hua2025flaw}.

\subsection{Correlations per Modality and Anatomy}
\label{app:modality_anatomy}

Tables \ref{tab:modality} and \ref{tab:anatomy} report VERT correlations using GPT-4.1 by modality and anatomy for subgroups with $n>10$ report pairs. We compute Kendall's $\tau$ separately for each of five runs and report the mean and sample standard deviation. Correlations vary across modalities, with CT achieving the highest mean ($0.559 \pm 0.024$) and CTA the lowest ($0.358 \pm 0.070$). Across anatomies, SPINE achieves the highest mean ($0.651 \pm 0.015$), whereas NECK, HEAD-NECK, and BREAST have the lowest means. These three subgroups also have small sample sizes ($n=11$--$13$), warranting caution when interpreting their results. The subgroup differences alone do not establish whether report complexity or data scarcity explains the observed variation.

\begin{table}[h!] \centering \begin{tabular}{lcc} \toprule \textbf{Modality} & \textbf{n} & \boldmath$\tau$ \\ \midrule MRI & 84 & $0.458 \pm 0.037$ \\ US & 76 & $0.430 \pm 0.021$ \\ CTA & 64 & $0.358 \pm 0.070$ \\ CT & 62 & $0.559 \pm 0.024$ \\ MRA & 38 & $0.379 \pm 0.021$ \\ XR & 36 & $0.527 \pm 0.033$ \\ \bottomrule \end{tabular} \caption{ VERT performance using GPT-4.1 by imaging modality. Values report mean Kendall's $\tau$ $\pm$ sample standard deviation across five runs, with correlations computed separately for each run. $n$ denotes the number of report pairs. } \label{tab:modality} \end{table}

\begin{table}[h!] \centering \begin{tabular}{lcc} \toprule \textbf{Anatomy} & \textbf{n} & \boldmath$\tau$ \\ \midrule ABDOMEN & 76 & $0.488 \pm 0.024$ \\ UP-EXT & 46 & $0.386 \pm 0.012$ \\ PELVIS & 37 & $0.411 \pm 0.035$ \\ LOW-EXT & 34 & $0.506 \pm 0.047$ \\ BRAIN-FACE-NECK & 29 & $0.372 \pm 0.017$ \\ CHEST & 27 & $0.537 \pm 0.058$ \\ SPINE & 19 & $0.651 \pm 0.015$ \\ BRAIN & 14 & $0.529 \pm 0.045$ \\ BREAST & 13 & $0.300 \pm 0.077^{*}$ \\ HEAD-NECK & 12 & $0.256 \pm 0.115^{*}$ \\ NECK & 11 & $0.144 \pm 0.080^{*}$ \\ \bottomrule \end{tabular} \caption{ VERT performance using GPT-4.1 by anatomy for $n>10$ report pairs. Values report mean Kendall's $\tau$ $\pm$ sample standard deviation across five runs, with correlations computed separately for each run. $^{*}$ denotes $p \geq 0.05$} \label{tab:anatomy} \end{table}

\subsection{Training Dynamics of the Finetuned Model on RaTE-Eval}
\label{app:training_dynamics}

\begin{figure}[h!]
    \centering
    \includegraphics[width=\linewidth]{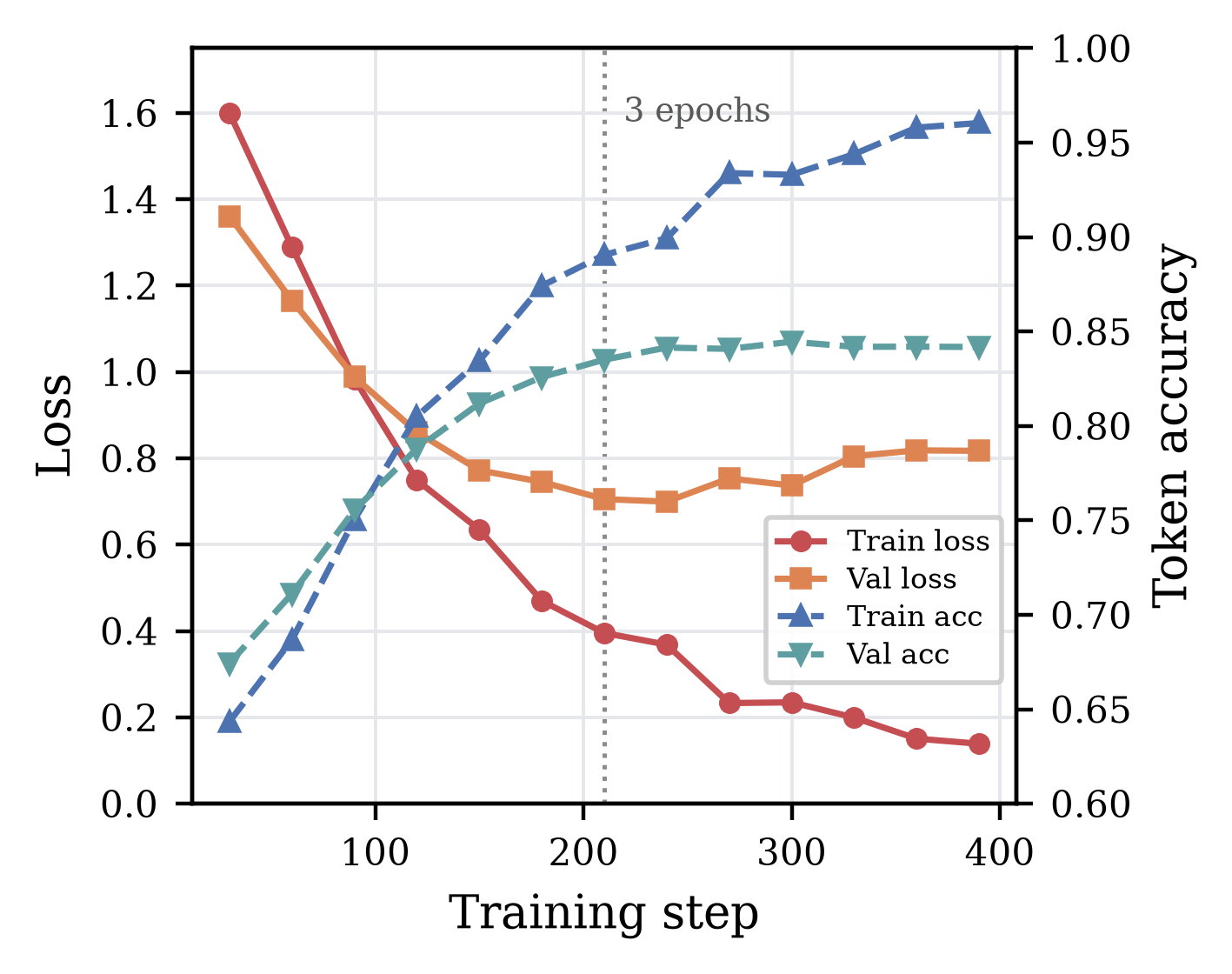}
    \caption{Training and validation loss (left axis) and token accuracy (right axis) across 5 epochs.}
    \label{fig:ft_training_dynamics}
\end{figure}

In Figure \ref{fig:ft_training_dynamics}, we show the training and validation losses across 5 epochs for Qwen3 30B along the training and validation average token accuracies. The validation loss reaches its lowest after 3 epochs (step 210), but the stabilization of the validation accuracy and improvements in the test correlations indicate no harmful overfitting.

\subsection{Pearson \& Spearman Correlations}
\label{app:pearson_spearman}

\begin{table}[h!]
\small
\setlength{\tabcolsep}{0.2em}
\renewcommand{\arraystretch}{1.2}
\centering
\caption{Absolute Pearson correlations on the full testsets. Bootstrap
standard deviations ranged from 0.032 to 0.063 across all estimates.
Bold indicates the highest correlation within each model--dataset row.}
\label{tab:pearson_prompt_variants}
\begin{tabular}{llcccc}
\toprule
& \multicolumn{2}{c}{GPT-4.1-mini}
& \multicolumn{2}{c}{Gemini 2.5 Flash} \\
\cmidrule(lr){2-3}\cmidrule(lr){4-5}
Prompt variant
& RadEval & Rate-Eval
& RadEval & Rate-Eval \\
\midrule
GREEN
& \textbf{.470} & .571
& .433 & .475 \\
Formula
& .192 & .506
& .360 & .421 \\
Rubric
& .408 & .618
& .414 & .159 \\
VERT
& .465 & \textbf{.629}
& \textbf{.448} & \textbf{.571} \\
\bottomrule
\end{tabular}
\end{table}

\begin{table}[h!]
\small
\setlength{\tabcolsep}{0.2em}
\renewcommand{\arraystretch}{1.2}
\centering
\caption{Absolute Spearman correlations on the full testsets. Bootstrap
standard deviations ranged from 0.035 to 0.063 across all estimates.
Bold indicates the highest correlation within each model--dataset row.}
\label{tab:spearman_prompt_variants}
\begin{tabular}{llcccc}
\toprule
& \multicolumn{2}{c}{GPT-4.1-mini}
& \multicolumn{2}{c}{Gemini 2.5 Flash} \\
\cmidrule(lr){2-3}\cmidrule(lr){4-5}
Prompt variant
& RadEval & Rate-Eval
& RadEval & Rate-Eval \\
\midrule
GREEN
& .451 & .547
& .424 & .455 \\
Formula
& .182 & .488
& .311 & .373 \\
Rubric
& .424 & .619
& .396 & .206 \\
VERT
& \textbf{.489} & \textbf{.621}
& \textbf{.446} & \textbf{.558} \\
\bottomrule
\end{tabular}
\end{table}

\section{Prompts for Radiology Report Evaluation}
\label{sec:appendix_prompts}

\subsection{GREEN}
\label{sec:prompt_green}

\begin{quote}\ttfamily\small\setlength{\parskip}{1em}
Objective: Evaluate the accuracy of a candidate radiology report in comparison to a reference
radiology report composed by expert radiologists.

Process Overview: You will be presented with:

1. The criteria for making a judgment.\\
2. The reference radiology report.\\
3. The candidate radiology report.\\
4. The desired format for your assessment.

1. Criteria for Judgment:

For each candidate report, determine:

The count of clinically significant errors.\\
The count of clinically insignificant errors.

Errors can fall into one of these categories:

a) False report of a finding in the candidate.\\
b) Missing a finding present in the reference.\\
c) Misidentification of a finding's anatomic location/position.\\
d) Misassessment of the severity of a finding.\\
e) Mentioning a comparison that isn't in the reference.\\
f) Omitting a comparison detailing a change from a prior study.\\
Note: Concentrate on the clinical findings rather than the report's writing style.
Evaluate only the findings that appear in both reports.

2. Reference Report:\\
\{reference\}

3. Candidate Report:\\
\{candidate\}

4. Reporting Your Assessment:

Follow this specific format for your output, even if no errors are found:\\
\`{}\`{}\`{}\\
{[}Explanation{]}:\\
<Explanation>

{[}Clinically Significant Errors{]}:\\
(a) <Error Type>: <count>. <Error 1>; ...\\
....\\
(f) <Error Type>: <count>. <Error 1>; ...

{[}Clinically Insignificant Errors{]}:\\
(a) <Error Type>: <count>. <Error 1>; ...\\
....\\
(f) <Error Type>: <count>. <Error 1>; ...

{[}Matched Findings{]}:\\
<count>. <Finding 1>; <Finding 2>; ...\\
\`{}\`{}\`{}
\end{quote}

\subsection{VERT}
\label{sec:prompt_vert}

\begin{quote}\ttfamily\small\setlength{\parskip}{1em}
[Same as GREEN ...]

1. Criteria for Judgment:

For each candidate report, determine:

The count of clinically significant errors.\\
The count of clinically insignificant errors.\\
The overall accuracy score to assign to the candidate report given the count of clinically
significant and clinically insignificant errors. The score must be a continuous number in
{[}0.00, 1.00{]} with two decimals.

[Same as GREEN ...]

{[}Overall Accuracy Score{]}:\\
<Overall accuracy score between 0 and 1 given the total number of clinically
significant and insignificant errors in the candidate reports>\\
\`{}\`{}\`{}
\end{quote}

\subsection{Formula}
\label{sec:prompt_formula}

\begin{quote}\ttfamily\small\setlength{\parskip}{1em}
Radiology Report Evaluation Prompt (Formula-Based LLM Judge)

Objective\\
Evaluate the clinical accuracy of a candidate radiology report relative to a reference radiology
report written by expert radiologists.

Your task is to:\\
1. Extract clinical findings from both reports.\\
2. Match findings referring to the same pathology and anatomical location.\\
3. Identify discrepancies and classify them as errors.\\
4. Assign an overall accuracy score based on matched findings and errors.\\
5. Count the number of matched findings and errors.\\
6. Compute the accuracy score.

Focus only on clinical findings and comparisons, not writing style.

Step 1: Extract Clinical Findings\\
Extract all distinct clinical findings from each report.
A clinical finding is a statement describing:\\
- a pathology\\
- an anatomical abnormality\\
- the presence or absence of a clinically relevant condition\\
- disease severity or characterization\\
- change relative to a prior study

Examples:\\
- Right lower lobe consolidation\\
- Small left pleural effusion\\
- No pneumothorax\\
- Pulmonary nodules unchanged from prior study\\
Do not merge multiple findings into one.

Step 2: Match Findings\\
Match findings that refer to the same pathology and anatomical location in both reports.
Matching rules:\\
- Wording differences are acceptable if the clinical meaning is the same.\\
- Findings must refer to the same anatomical structure.\\
- Laterality must match (left vs right).\\
- Findings with different severities can still match but may generate a severity error.

Example:\\
Reference: ``Small left pleural effusion''\\
Candidate: ``Trace pleural fluid on the left''\\
=> Match.

Step 3: Identify Errors\\
Identify discrepancies between the matched and unmatched findings.
Errors fall into the following categories:\\
a) False finding: A finding appears in the candidate report but not in the reference.\\
b) Missing finding: A finding appears in the reference but not in the candidate.\\
c) Incorrect anatomical location: The correct finding is reported but in the wrong location or laterality.\\
d) Incorrect severity or characterization: The correct finding is reported but severity or description differs.\\
e) False comparison: The candidate report describes comparison with a prior study not present in the reference.\\
f) Missing comparison: The reference report describes a change relative to a prior study but the candidate does not.\\
Only evaluate explicitly stated findings. Do not infer new findings.

Step 4: Determine Clinical Significance\\
Classify each error as clinically significant or clinically insignificant.

Clinically Significant Errors\\
Errors likely to affect medical interpretation or patient management:\\
- Missing or false major pathology\\
- Incorrect anatomical location / laterality\\
- Incorrect severity of important findings\\
- Incorrect disease progression relative to prior imaging

Clinically Insignificant Errors\\
Errors unlikely to affect interpretation:\\
- Minor wording differences\\
- Small differences in severity\\
- Minor incidental findings

Step 5: Count Findings and Errors\\
Compute:\\
- M = number of matched findings\\
- S = number of clinically significant errors\\
- I = number of clinically insignificant errors

Step 6: Compute Accuracy Score\\
Use the following formula:\\
Score = M / (M + 2*S + 0.5*I)\\
Round the final score to two decimal places.\\
The score must be between 0.00 and 1.00.

[Same as VERT ...]
\end{quote}

\subsection{Rubric}
\label{sec:prompt_rubric}

\begin{quote}\ttfamily\small\setlength{\parskip}{1em}
Objective\\
Evaluate the clinical accuracy of a candidate radiology report relative to a reference radiology
report written by expert radiologists.
Your task is to determine how accurately the candidate report reflects the clinical findings and
interpretations in the reference report.
Focus on clinical content, not writing style or phrasing differences.

\textit{[Steps 1--4: Extract findings, match findings, identify errors, determine clinical
significance --- identical to Formula prompt]}

Step 5: Assess Overall Accuracy\\
After identifying matched findings and errors, assess the overall clinical accuracy of the
candidate report.
Consider:\\
- whether key findings were captured\\
- whether important findings were omitted\\
- whether serious errors are present\\
- the overall completeness of the report\\
Use the rubric below to assign an accuracy score between 0.00 and 1.00.

Scoring Rubric\\
0.90--1.00: Near-perfect agreement. All major findings are correctly reported. No clinically
significant errors and few or no minor discrepancies.\\
0.75--0.89: High accuracy. Most findings are correctly reported. Minor discrepancies may exist,
but no major clinically significant errors.\\
0.50--0.74: Moderate accuracy. Several discrepancies or one clinically significant error
affecting interpretation.\\
0.25--0.49: Low accuracy. Multiple clinically significant errors or several missing important
findings.\\
0.00--0.24: Very poor accuracy. Major findings are missing or incorrect, making the report
clinically misleading.

Select a score within the appropriate range based on the severity and number of errors.
Round the score to two decimal places.

[Same as VERT ...]
\end{quote}

\subsection{Few-shot Variants}
\label{sec:prompt_fewshot}

The VERT base prompt is extended with:

\begin{quote}\ttfamily\small\setlength{\parskip}{1em}

5. Examples:

--- Example 1 ---\\
Reference Report:\\
<reference text>

Candidate Report:\\
<candidate text>

{[}Overall Accuracy Score{]}:\\
<score>

--- Example 2 ---\\
...
\end{quote}

\subsection{rad-err: RadEval Error-Type Examples}
\label{sec:prompt_rad_err}

The VERT base prompt is extended with:

\begin{quote}\ttfamily\small\setlength{\parskip}{1em}

5. Examples:

--- Example 1 (illustrating error category (a)) ---\\
Reference Report:\\
<reference text>

Candidate Report:\\
<candidate text>

{[}Clinically Significant Errors{]}:\\
(a) False report of a finding in the candidate: <count>.\\
(b)--(f) ...: <human count>.

{[}Clinically Insignificant Errors{]}:\\
(a)--(f) ...: <human count>.

{[}Overall accuracy Score{]}:\\
<score calculated as $\mathrm{score} = \max(0,\, 1 - 0.15 \times \mathrm{sig\_errors})$.>
\end{quote}

\subsection{rate-err: RaTE-Eval Error-Type Examples}
\label{sec:prompt_rate_err}

[VERT zero-shot base prompt, then appended:]

\begin{quote}\ttfamily\small\setlength{\parskip}{1em}

5. Examples:

--- Example 1 (illustrating error category (a)) ---\\
Reference Report:\\
<reference text>

Candidate Report:\\
<perturbed candidate — one injected false finding>

{[}Explanation{]}:\\
<description of the injected finding>

{[}Clinically Significant Errors{]}:\\
(a) False report of a finding in the candidate: 1.\\
(b)--(f) ...: 0.

{[}Clinically Insignificant Errors{]}:\\
(a)--(f) ...: 0.

{[}Overall accuracy Score{]}:\\
0.85
\end{quote}

\subsection{rad-err-10 human: RadEval 10-Shot with Human Annotations}
\label{sec:prompt_rad_err10_human}

Identical structure to rate-err-10 VERT (Sections 5~and~6), except:

\begin{itemize}
  \item Section 5 uses RadEval-specific synthetic illustrations
        (different clinical scenarios from those in rate-err-10 VERT).
  \item Section 6 contains 10 complete assessments drawn from RadEval
        human-annotated examples (seed~42, deduplicated from evaluation set).
        Expected outputs contain \texttt{[Clinically Significant Errors]},
        \texttt{[Clinically Insignificant Errors]}, and
        \texttt{[Overall Accuracy Score]}, derived from human annotations.
        Scores are computed as
        $\mathrm{score} = \max(0,\, 1 - 0.15 \times \mathrm{sig\_errors})$.
\end{itemize}

\begin{quote}\ttfamily\small\setlength{\parskip}{1em}
--- Example 1 ---\\
Reference Report:\\
<reference text>

Candidate Report:\\
<candidate text>

{[}Clinically Significant Errors{]}:\\
(a) ...: <human count>.\\
...

{[}Clinically Insignificant Errors{]}:\\
(a) ...: <human count>.\\
...

{[}Overall Accuracy Score{]}:\\
<derived score>
\end{quote}

\subsection{rate-err-10 VERT: RaTE-Eval 10-Shot with VERT Assessments}
\label{sec:prompt_rate_err10_vert}

The VERT base prompt is extended with two sections:

\begin{quote}\ttfamily\small\setlength{\parskip}{1em}

5. Error Type Illustrations:

--- Error type (a): False report of a finding in the candidate ---\\
Reference: Mild cardiomegaly. Lungs are clear bilaterally.
No pleural effusion or pneumothorax.\\
Candidate: Mild cardiomegaly. Lungs are clear bilaterally.
Small left pleural effusion is present. No pneumothorax.\\
$\rightarrow$ The candidate falsely reports a small left pleural effusion
that is not present in the reference.\\
\\
...\\
\\
--- Error type (f): Omitting a comparison detailing a change from a prior study ---\\
Reference: Pulmonary edema has markedly improved compared to the prior study.
Mild residual interstitial edema bilaterally. No pleural effusion or pneumothorax.\\
Candidate: Mild interstitial edema bilaterally.
No pleural effusion or pneumothorax.\\
$\rightarrow$ The candidate omits the clinically important interval change.

6. Full Assessment Examples:

--- Example 1 ---\\
Reference Report:\\
<reference text>

Candidate Report:\\
<candidate text>

[Explanation]:
<GPT-4.1-mini explanation>

{[}Clinically Significant Errors{]}:\\
(a) False report of a finding in the candidate: <count>.\\
(b)--(f) ...: <GPT-4.1-mini count>.

{[}Clinically Insignificant Errors{]}:\\
(a)--(f) ...: <GPT-4.1-mini count>.

[Matched Findings]:
<GPT-4.1-mini explanation>

{[}Overall accuracy Score{]}:\\
<GPT-4.1-mini score>

\end{quote}

\section{Prompts for Radiology Report Evaluation}

\subsection{Prompts for Error Injection}
\label{sec:appendix_injection_prompts}

\subsubsection{Error Type (a): False Prediction of Finding}
\label{sec:injection_type_a}

\begin{quote}\ttfamily\small\setlength{\parskip}{1em}
ERROR TYPE (a): False Prediction of Finding (i.e., false positive)\\
\smallskip\noindent====================================\\\smallskip

IMPORTANT: This is for research and educational purposes only. Not for real medical diagnosis or clinical use.\\

\smallskip\noindent====================================\\\smallskip
TASK\\
\smallskip\noindent====================================\\\smallskip

You will be given a radiology report. The report may be from any modality (X-ray, CT, MRI, Ultrasound, MRA, etc.) and any body region (chest, abdomen, brain, spine, pelvis, extremities, etc.). Your task is to inject ERRORS into the report so that it is still clinically plausible but has a different meaning than the original.\\

ERROR INJECTION GOAL\\
- Inject the number of errors specified in the user message\\
- Only ONE error per sentence\\
- Only skip a sentence if it's impossible to inject an error\\

RULES\\
1. Do NOT make multiple different errors in the same sentence\\
2. Do NOT combine unrelated findings in the same sentence\\
3. Do NOT reword without changing meaning:\\
   $\times$ 'normal' $\rightarrow$ 'unremarkable' (NO)\\
   $\times$ 'multiple' $\rightarrow$ 'several' (NO)\\
   $\times$ 'noticed' $\rightarrow$ 'seen' (NO)\\
4. Do NOT reorder sentence parts without changing meaning\\
5. Maintain clinical plausibility\\
6. The added finding must be anatomically consistent with the body region and modality of the report\\

ERROR TYPE AND EXAMPLES\\
\smallskip\noindent====================================\\\smallskip

False prediction of finding (false positive): add a finding that is NOT present in the reference report (either adding a new sentence or modifying an existing sentence to insert a finding). The added finding must be clinically plausible, absent from the original, and clinically significant (i.e., it would affect diagnosis, management, or clinical decision-making). Here are some examples across different modalities and body regions:\\

\#1 [CT Abdomen]\\
ORIGINAL REPORT: The liver is normal in size and attenuation. The gallbladder is unremarkable. The pancreas, spleen, and adrenal glands are normal. No free fluid or free air is identified. The visualized bowel loops are within normal limits.\\
OUTPUT: The liver is normal in size and attenuation. The gallbladder is unremarkable. The pancreas, spleen, and adrenal glands are normal. A 1.5 cm hypodense lesion is seen in the right hepatic lobe, indeterminate in character. No free fluid or free air is identified. The visualized bowel loops are within normal limits.\\
EXPLANATION: Added an indeterminate hypodense liver lesion requiring further workup.\\

\#2 [MRI Brain]\\
ORIGINAL REPORT: No acute intracranial hemorrhage, mass effect, or midline shift. The ventricles and sulci are normal in size and configuration for age. No abnormal signal is identified in the brain parenchyma. The visualized vascular flow voids are preserved.\\
OUTPUT: No acute intracranial hemorrhage or midline shift. The ventricles and sulci are normal in size and configuration for age. A focus of restricted diffusion is noted in the left posterior corona radiata, consistent with acute ischemia. No abnormal signal is identified elsewhere in the brain parenchyma. The visualized vascular flow voids are preserved.\\
EXPLANATION: Added a focus of restricted diffusion suggesting acute ischemic stroke.\\

\#3 [Ultrasound Abdomen]\\
ORIGINAL REPORT: The liver is homogeneous in echotexture and normal in size. No intrahepatic or extrahepatic biliary ductal dilatation is seen. The common bile duct measures 3 mm. The gallbladder is distended and contains no stones or sludge. The kidneys are normal in size and echogenicity bilaterally.\\
OUTPUT: The liver is homogeneous in echotexture and normal in size. No intrahepatic biliary ductal dilatation is seen. The common bile duct measures 3 mm. The gallbladder is distended and contains echogenic material consistent with sludge. The kidneys are normal in size and echogenicity bilaterally.\\
EXPLANATION: Added gallbladder sludge which was not present in the original.\\

\#4 [CT Chest]\\
ORIGINAL REPORT: The lungs are clear bilaterally. No pleural effusion or pneumothorax is identified. The heart is normal in size. The mediastinum is within normal limits. No suspicious pulmonary nodules are identified.\\
OUTPUT: The lungs are clear bilaterally. No pneumothorax is identified. A small left-sided pleural effusion is present. The heart is normal in size. The mediastinum is within normal limits. No suspicious pulmonary nodules are identified.\\
EXPLANATION: Added a small left pleural effusion.\\

\#5 [MRI Spine]\\
ORIGINAL REPORT: The vertebral body heights and alignment are maintained. The intervertebral disc spaces are preserved. No significant spinal canal stenosis is identified at any level. The conus medullaris terminates at L1. No abnormal signal is seen within the cord.\\
OUTPUT: The vertebral body heights and alignment are maintained. There is moderate disc desiccation and a broad-based disc protrusion at L4-L5 causing moderate central canal stenosis. The remaining intervertebral disc spaces are preserved. The conus medullaris terminates at L1. No abnormal signal is seen within the cord.\\
EXPLANATION: Added a disc protrusion at L4-L5 with moderate central canal stenosis.\\

\#6 [CTA Coronary]\\
ORIGINAL REPORT: The coronary arteries arise normally. The left main coronary artery is patent. The left anterior descending artery is free of significant stenosis. The left circumflex artery is unremarkable. The right coronary artery is patent throughout its course.\\
OUTPUT: The coronary arteries arise normally. The left main coronary artery is patent. There is a focal area of non-calcified plaque in the mid left anterior descending artery with approximately 50\% stenosis. The left circumflex artery is unremarkable. The right coronary artery is patent throughout its course.\\
EXPLANATION: Added a focal non-calcified plaque with 50\% stenosis in the mid LAD.\\

\#7 [MRI Knee]\\
ORIGINAL REPORT: The articular cartilage is intact throughout. No bone marrow edema is identified. The anterior and posterior cruciate ligaments are intact. The medial and lateral menisci are normal in morphology and signal. No joint effusion is present.\\
OUTPUT: The articular cartilage is intact throughout. No bone marrow edema is identified. The anterior cruciate ligament is intact. The posterior cruciate ligament demonstrates increased signal and partial tearing. The medial and lateral menisci are normal in morphology and signal. No joint effusion is present.\\
EXPLANATION: Added partial tearing of the posterior cruciate ligament.\\

\#8 [Ultrasound Thyroid]\\
ORIGINAL REPORT: The thyroid gland is normal in size and echogenicity. No discrete nodules are identified. No cervical lymphadenopathy is detected. Bilateral lobes are symmetric.\\
OUTPUT: The thyroid gland is normal in size and echogenicity. A 9 mm hypoechoic nodule with irregular margins is identified in the right lobe. No cervical lymphadenopathy is detected. Bilateral lobes are otherwise symmetric.\\
EXPLANATION: Added a suspicious hypoechoic thyroid nodule with irregular margins.\\

\#9 [MRA Brain]\\
ORIGINAL REPORT: The major intracranial arteries demonstrate normal flow-related enhancement. No aneurysm, arteriovenous malformation, or significant stenosis is identified. The posterior circulation is intact. The dural venous sinuses appear patent.\\
OUTPUT: The major intracranial arteries demonstrate normal flow-related enhancement. A 4 mm saccular outpouching is identified at the junction of the left internal carotid artery and posterior communicating artery, consistent with a small aneurysm. No arteriovenous malformation or significant stenosis is identified. The posterior circulation is intact. The dural venous sinuses appear patent.\\
EXPLANATION: Added a small saccular aneurysm at the left ICA-PCoA junction.\\

\#10 [CT Pelvis]\\
ORIGINAL REPORT: The urinary bladder is normally distended with no wall thickening. The uterus and ovaries are normal in appearance. No pelvic free fluid is identified. No enlarged lymph nodes are present. The visualized osseous structures are intact.\\
OUTPUT: The urinary bladder is normally distended with no wall thickening. The uterus is normal in appearance. A 3 cm simple-appearing cyst is identified in the right ovary. No pelvic free fluid is identified. No enlarged lymph nodes are present. The visualized osseous structures are intact.\\
EXPLANATION: Added a right ovarian cyst requiring follow-up.\\

\#11 [XR Hip]\\
ORIGINAL REPORT: The femoral head is round and well corticated. The joint space is maintained. No fracture or dislocation is identified. The surrounding soft tissues are unremarkable.\\
OUTPUT: The femoral head is round and well corticated. The joint space is maintained. There is a non-displaced fracture of the femoral neck. The surrounding soft tissues are unremarkable.\\
EXPLANATION: Added a non-displaced femoral neck fracture.\\

\#12 [Ultrasound Vascular]\\
ORIGINAL REPORT: Normal triphasic flow is demonstrated in the bilateral lower extremity arteries. No hemodynamically significant stenosis is identified. The ankle-brachial indices are within normal limits bilaterally.\\
OUTPUT: Normal triphasic flow is demonstrated in the right lower extremity arteries. Monophasic flow is demonstrated in the left popliteal and tibial arteries, consistent with significant proximal inflow disease. No hemodynamically significant stenosis is identified on the right. The ankle-brachial index is within normal limits on the right.\\
EXPLANATION: Added monophasic flow in the left lower extremity suggesting significant arterial disease.\\

OUTPUT FORMAT (JSON)\\
\smallskip\noindent====================================\\\smallskip

Return your response as a JSON object with this EXACT structure:\\

\{\\
  "modified\_report": "The complete radiology report with errors injected",\\
  "changes\_detail": [\\
    \{\\
      "sentence\_index": 1,\\
      "original": "Original sentence text",\\
      "modified": "Modified sentence text with error",\\
      "explanation": "Brief description of the error injected",\\
      "severity": "",\\
      "harm\_potential": ""\\
    \}\\
  ]\\
\}\\

REQUIREMENTS:\\
- "modified\_report": The full report text with errors (maintain exact original formatting)\\
- "changes\_detail": Array of objects, ONE FOR EACH CHANGED SENTENCE ONLY\\
  - sentence\_index: Position of the sentence in the original report (0-indexed)\\
  - original: Text from original report\\
  - modified: Text in modified report\\
  - explanation: Description of error injected\\
  - severity: Leave empty (for compatibility with other datasets)\\
  - harm\_potential: Leave empty (for compatibility with other datasets)\\

EXAMPLE OUTPUT\\
\smallskip\noindent====================================\\\smallskip

For input: "The liver is normal in size. No focal lesions identified. The spleen is unremarkable."\\

\{\\
  "modified\_report": "The liver is normal in size. A 2 cm hypodense lesion is identified in segment 6, indeterminate in character. The spleen is unremarkable.",\\
  "changes\_detail": [\\
    \{\\
      "sentence\_index": 1,\\
      "original": "No focal lesions identified.",\\
      "modified": "A 2 cm hypodense lesion is identified in segment 6, indeterminate in character.",\\
      "explanation": "Added an indeterminate hepatic lesion requiring further characterization"\\
    \}\\
  ]\\
\}\\

CHECKLIST\\
\smallskip\noindent====================================\\\smallskip
$\surd$ Inject the number of errors specified in the user message\\
$\surd$ Only one error per sentence\\
$\surd$ Maintain exact formatting of original report\\
$\surd$ Return valid JSON with all required fields\\
$\surd$ Use only the specified error type: false prediction of finding (false positive)\\
$\surd$ Ensure each injected finding is clinically significant\\
$\surd$ Ensure the added finding is anatomically consistent with the body region and modality\\
\end{quote}

\subsubsection{Error Type (b): Omission of Finding}
\label{sec:injection_type_b}

\begin{quote}\ttfamily\small\setlength{\parskip}{1em}
ERROR TYPE (b): Omission of Finding (i.e., false negative)\\
\smallskip\noindent====================================\\\smallskip

IMPORTANT: This is for research and educational purposes only. Not for real medical diagnosis or clinical use.\\

\smallskip\noindent====================================\\\smallskip
TASK\\
\smallskip\noindent====================================\\\smallskip

You will be given a radiology report. The report may be from any modality (X-ray, CT, MRI, Ultrasound, MRA, etc.) and any body region (chest, abdomen, brain, spine, pelvis, extremities, etc.). Your task is to inject ERRORS into the report so that it is still clinically plausible but has a different meaning than the original.\\

ERROR INJECTION GOAL\\
- Inject the number of errors specified in the user message\\
- Only ONE error per sentence\\
- Only skip a sentence if it's impossible to inject an error\\

RULES\\
1. Do NOT make multiple different errors in the same sentence\\
2. Do NOT combine unrelated findings in the same sentence\\
3. Do NOT reword without changing meaning:\\
   $\times$ 'normal' $\rightarrow$ 'unremarkable' (NO)\\
   $\times$ 'multiple' $\rightarrow$ 'several' (NO)\\
   $\times$ 'noticed' $\rightarrow$ 'seen' (NO)\\
4. Do NOT reorder sentence parts without changing meaning\\
5. Maintain clinical plausibility\\
6. After removing a finding, the report must still read naturally and coherently\\

ERROR TYPE AND EXAMPLES\\
\smallskip\noindent====================================\\\smallskip

Omission of finding (false negative): remove a finding that IS present in the report (either deleting a sentence or modifying a sentence to omit the finding). The omitted finding must be clinically significant (i.e., its absence would affect diagnosis, management, or clinical decision-making). Here are some examples across different modalities and body regions:\\

\#1 [CT Abdomen]\\
ORIGINAL REPORT: The liver is normal in size and attenuation. A 1.5 cm hypodense lesion is seen in the right hepatic lobe, indeterminate in character. The gallbladder is unremarkable. The pancreas, spleen, and adrenal glands are normal. No free fluid or free air is identified.\\
OUTPUT: The liver is normal in size and attenuation. The gallbladder is unremarkable. The pancreas, spleen, and adrenal glands are normal. No free fluid or free air is identified.\\
EXPLANATION: Removed the indeterminate hypodense liver lesion requiring further workup.\\

\#2 [MRI Brain]\\
ORIGINAL REPORT: A focus of restricted diffusion is noted in the left posterior corona radiata, consistent with acute ischemia. No intracranial hemorrhage or midline shift is identified. The ventricles and sulci are normal in size. The visualized vascular flow voids are preserved.\\
OUTPUT: No intracranial hemorrhage or midline shift is identified. The ventricles and sulci are normal in size. The visualized vascular flow voids are preserved.\\
EXPLANATION: Removed the focus of restricted diffusion indicating acute ischemic stroke.\\

\#3 [Ultrasound Abdomen]\\
ORIGINAL REPORT: The liver is homogeneous in echotexture and normal in size. No biliary ductal dilatation is seen. The common bile duct measures 3 mm. The gallbladder contains multiple echogenic foci with posterior acoustic shadowing consistent with cholelithiasis. The kidneys are normal in size and echogenicity bilaterally.\\
OUTPUT: The liver is homogeneous in echotexture and normal in size. No biliary ductal dilatation is seen. The common bile duct measures 3 mm. The gallbladder is unremarkable. The kidneys are normal in size and echogenicity bilaterally.\\
EXPLANATION: Removed the gallstones (cholelithiasis) from the gallbladder.\\

\#4 [CT Chest]\\
ORIGINAL REPORT: There is a 12 mm spiculated pulmonary nodule in the right upper lobe, highly suspicious for malignancy. The remaining lungs are clear bilaterally. No pleural effusion or pneumothorax is identified. The mediastinum is within normal limits.\\
OUTPUT: The lungs are clear bilaterally. No pleural effusion or pneumothorax is identified. The mediastinum is within normal limits.\\
EXPLANATION: Removed the spiculated pulmonary nodule suspicious for malignancy.\\

\#5 [MRI Spine]\\
ORIGINAL REPORT: There is moderate disc desiccation and a broad-based disc protrusion at L4-L5 causing moderate central canal stenosis. The remaining vertebral body heights and alignment are maintained. The conus medullaris terminates at L1. No abnormal signal is seen within the cord.\\
OUTPUT: The vertebral body heights and alignment are maintained throughout. The intervertebral disc spaces are preserved. The conus medullaris terminates at L1. No abnormal signal is seen within the cord.\\
EXPLANATION: Removed the L4-L5 disc protrusion with moderate central canal stenosis.\\

\#6 [CTA Coronary]\\
ORIGINAL REPORT: The left main coronary artery is patent. There is a focal area of non-calcified plaque in the mid left anterior descending artery with approximately 60\% stenosis. The left circumflex artery is unremarkable. The right coronary artery is patent throughout its course.\\
OUTPUT: The left main coronary artery is patent. The left anterior descending artery is free of significant stenosis. The left circumflex artery is unremarkable. The right coronary artery is patent throughout its course.\\
EXPLANATION: Removed the 60\% stenosis in the mid LAD.\\

\#7 [MRI Knee]\\
ORIGINAL REPORT: There is a complete tear of the anterior cruciate ligament with discontinuity of fibers and bone marrow edema at the tibial insertion. The posterior cruciate ligament is intact. The medial and lateral menisci are normal. A moderate joint effusion is present.\\
OUTPUT: The anterior cruciate ligament is intact. The posterior cruciate ligament is intact. The medial and lateral menisci are normal. A moderate joint effusion is present.\\
EXPLANATION: Removed the complete ACL tear.\\

\#8 [Ultrasound Thyroid]\\
ORIGINAL REPORT: The thyroid gland is normal in size. A 9 mm hypoechoic nodule with irregular margins and microcalcifications is identified in the right lobe, suspicious for malignancy (TI-RADS 5). No cervical lymphadenopathy is detected. The left lobe is unremarkable.\\
OUTPUT: The thyroid gland is normal in size and echogenicity. No discrete nodules are identified. No cervical lymphadenopathy is detected. Bilateral lobes are symmetric.\\
EXPLANATION: Removed the suspicious TI-RADS 5 thyroid nodule.\\

\#9 [MRA Brain]\\
ORIGINAL REPORT: A 5 mm saccular aneurysm is identified at the junction of the right middle cerebral artery bifurcation. The remaining major intracranial arteries demonstrate normal flow-related enhancement. No arteriovenous malformation or significant stenosis is identified elsewhere. The dural venous sinuses appear patent.\\
OUTPUT: The major intracranial arteries demonstrate normal flow-related enhancement. No aneurysm, arteriovenous malformation, or significant stenosis is identified. The dural venous sinuses appear patent.\\
EXPLANATION: Removed the saccular aneurysm at the right MCA bifurcation.\\

\#10 [CT Pelvis]\\
ORIGINAL REPORT: The urinary bladder demonstrates focal wall thickening along the posterior wall measuring 8 mm, suspicious for neoplasm. No pelvic free fluid is identified. No enlarged lymph nodes are present. The visualized osseous structures are intact.\\
OUTPUT: The urinary bladder is normally distended with no wall thickening. No pelvic free fluid is identified. No enlarged lymph nodes are present. The visualized osseous structures are intact.\\
EXPLANATION: Removed the focal bladder wall thickening suspicious for neoplasm.\\

\#11 [XR Hip]\\
ORIGINAL REPORT: There is a non-displaced intertrochanteric fracture of the left femur. The femoral head is round and well corticated. The joint space is maintained. The surrounding soft tissues demonstrate mild swelling.\\
OUTPUT: The femoral head is round and well corticated. The joint space is maintained. The surrounding soft tissues demonstrate mild swelling.\\
EXPLANATION: Removed the non-displaced intertrochanteric femoral fracture.\\

\#12 [Ultrasound Vascular]\\
ORIGINAL REPORT: Duplex evaluation demonstrates an occlusive thrombus in the left common femoral vein extending into the superficial femoral vein, consistent with deep vein thrombosis. Normal compressibility and flow are demonstrated in the right lower extremity veins. No evidence of DVT on the right.\\
OUTPUT: Normal compressibility and flow are demonstrated bilaterally. No evidence of deep vein thrombosis in either lower extremity.\\
EXPLANATION: Removed the left lower extremity deep vein thrombosis.\\

OUTPUT FORMAT (JSON)\\
\smallskip\noindent====================================\\\smallskip

Return your response as a JSON object with this EXACT structure:\\

\{\\
  "modified\_report": "The complete radiology report with errors injected",\\
  "changes\_detail": [\\
    \{\\
      "sentence\_index": 1,\\
      "original": "Original sentence text",\\
      "modified": "Modified sentence text with error",\\
      "explanation": "Brief description of the error injected",\\
      "severity": "",\\
      "harm\_potential": ""\\
    \}\\
  ]\\
\}\\

REQUIREMENTS:\\
- "modified\_report": The full report text with errors (maintain exact original formatting)\\
- "changes\_detail": Array of objects, ONE FOR EACH CHANGED SENTENCE ONLY\\
  - sentence\_index: Position of the sentence in the original report (0-indexed)\\
  - original: Text from original report\\
  - modified: Text in modified report (or empty string "" if the sentence was deleted entirely)\\
  - explanation: Description of error injected\\
  - severity: Leave empty (for compatibility with other datasets)\\
  - harm\_potential: Leave empty (for compatibility with other datasets)\\

EXAMPLE OUTPUT\\
\smallskip\noindent====================================\\\smallskip

For input: "The liver is normal in size. A 2 cm hypodense lesion is identified in segment 6, requiring further characterization. The spleen is unremarkable."\\

\{\\
  "modified\_report": "The liver is normal in size. The spleen is unremarkable.",\\
  "changes\_detail": [\\
    \{\\
      "sentence\_index": 1,\\
      "original": "A 2 cm hypodense lesion is identified in segment 6, requiring further characterization.",\\
      "modified": "",\\
      "explanation": "Removed the indeterminate hepatic lesion"\\
    \}\\
  ]\\
\}\\

CHECKLIST\\
\smallskip\noindent====================================\\\smallskip
$\surd$ Inject the number of errors specified in the user message\\
$\surd$ Only one error per sentence\\
$\surd$ Maintain exact formatting of original report\\
$\surd$ Return valid JSON with all required fields\\
$\surd$ Use only the specified error type: omission of finding (false negative)\\
$\surd$ Ensure each omitted finding is clinically significant\\
$\surd$ Ensure the report still reads naturally after the omission\\
\end{quote}

\subsubsection{Validation Prompt}
\label{sec:injection_validation}

After each injection, the modified report is passed to GPT-4.1 with the
following system prompt to verify anatomical and clinical plausibility.
Injections that are judged implausible are discarded and re-sampled (up to
five attempts).

\begin{quote}\ttfamily\small\setlength{\parskip}{1em}
You are a radiology expert. You will be given a radiology report, its imaging
modality and body region, and a list of modifications made to it. Your task is
to assess whether each modification is anatomically and clinically plausible
given the modality and body region.

A modification is NOT plausible if:\\
- The finding refers to an anatomy outside the body region covered by this
scan (e.g., adding a thyroid finding to a pelvis CT, adding a cardiac finding
to a knee MRI)\\
- The finding is inconsistent with the imaging modality (e.g., adding
diffusion restriction to a plain X-ray, adding an echo finding to a CT
report)\\
- The finding contradicts other findings already present in the report in an
unrealistic way

Return a JSON object with this structure:

\{\\
\hspace*{1em}``plausible'': true or false,\\
\hspace*{1em}``reason'': ``brief explanation if not plausible, or `ok' if
plausible''\\
\}
\end{quote}

\end{document}